\documentclass[10pt, conference, letterpaper]{IEEEtran}

\usepackage{soul}
\usepackage[dvipsnames]{xcolor}
\usepackage{cite}
\usepackage{setspace} 
\usepackage{subcaption}
\ifCLASSINFOpdf
  \usepackage[pdftex]{graphicx}
  \usepackage[underline=true,rounded corners=false]{pgf-umlsd}
\else
  \usepackage[dvips]{graphicx}
\fi
\usepackage{mathtools}
\usepackage{amssymb}
\usepackage[cal=cm]{mathalfa}
\usepackage{url}
\usepackage{listings}
\usepackage[noend]{algpseudocode}
\usepackage{amssymb, amsmath}
\usepackage{amsthm}
\usepackage{nccmath}
\usepackage[Algorithm]{algorithm}
\usepackage{comment}
\usepackage{svg}
\usepackage{soul}
\usepackage{afterpage}
\usepackage{cleveref}
\usepackage{booktabs}
\usepackage{amsfonts}
\usepackage{soul}
\usepackage{cancel}
\usepackage{balance}

\def\BibTeX{{\rm B\kern-.05em{\sc i\kern-.025em b}\kern-.08em
    T\kern-.1667em\lower.7ex\hbox{E}\kern-.125emX}}

\newcommand{\keepcomment}{0} 
\newcommand{\isrevision}{0} 

\newcommand{\algname}[0]{ASET}
\newcommand{\algnamelower}[0]{Aset}
\newcommand{\algfullname}[0]{Adaptive Scheduling of Edge Tasks}

\algnewcommand{\IfThen}[2]{
  \State \algorithmicif\ #1\ \algorithmicthen\ #2}

\usepackage[normalem]{ulem}  

\usepackage{xspace}

\ifnum\keepcomment=1
	\newcommand{\fulvio}[1]{{\leavevmode\color{red}{[FUL: #1]}}}
	\newcommand{\gabriele}[1]{{\leavevmode\color{DarkOrchid}{[GAB: #1]}}}
	\newcommand{\luq}[1]{{\leavevmode\color{green}{[luq: #1]}}}
	\newcommand{\carlos}[1]{{\leavevmode\color{PineGreen}{[carlos: #1]}}}
	\newcommand{\fda}[1]{{\leavevmode\color{brown}{[ferran: #1]}}}
	\newcommand{\note}[1]{\noindent\textcolor{gray}{#1}}
	\newcommand{\delete}[1]{\leavevmode\sout{#1}}
    
    \newcommand{\gc}[1]{\textcolor{DarkOrchid}{#1}}
    \newcommand{\ar}[1]{{\leavevmode\color{blue}{[ar: #1]}}}
	\newcommand{\jj}[1]{\textcolor{orange}{#1}}
	\newcommand{\gcdelete}[1]{\textcolor{DarkOrchid}{\leavevmode\sout{#1}}}
	\newcommand{\jjdelete}[1]{\textcolor{orange}{\leavevmode\sout{#1}}}
\else
    \newcommand{\fulvio}[1]{\leavevmode\ignorespaces\unskip}
	\newcommand{\gabriele}[1]{\leavevmode\ignorespaces\unskip}
	\newcommand{\luq}[1]{\leavevmode\ignorespaces\unskip}
	\newcommand{\carlos}[1]{\leavevmode\ignorespaces\unskip}
	\newcommand{\fda}[1]{\leavevmode\ignorespaces\unskip}
	\newcommand{\ar}[1]{\ignorespaces#1\ignorespaces\unskip}
    
    \newcommand{\gc}[1]{\ignorespaces#1\ignorespaces\unskip}
    \newcommand{\jj}[1]{\ignorespaces#1\ignorespaces\unskip}
    \newcommand{\note}[1]{\leavevmode\ignorespaces\unskip}
	\newcommand{\delete}[1]{\leavevmode\ignorespaces\unskip}
	\newcommand{\gcdelete}[1]{\leavevmode\ignorespaces\unskip}
	\newcommand{\jjdelete}[1]{\leavevmode\ignorespaces\unskip}
	\renewcommand{\cancel}[1]{\leavevmode\ignorespaces\unskip}
\fi

\ifnum\isrevision=1
    
\else
    
\fi

\ifnum\keepcomment=1
    \newcounter{DPNumberOfComments}
    \stepcounter{DPNumberOfComments}
    \newcommand{\dpnote}[1]{\textcolor{ForestGreen}{\small \bf [DP\#\arabic{DPNumberOfComments}\stepcounter{DPNumberOfComments}: #1]}}

    \newcommand{\NOTE}[1]
    {
      {\footnotesize\it
        \begin{center}
          \begin{tabular}{|c|}
           \hline
            \parbox{0.85\columnwidth}{
              \medskip
              #1
              \medskip} \\
            \hline
          \end{tabular}
        \end{center}
        }
    }
\else
    \newcommand\dpnote[1]{\leavevmode\ignorespaces\unskip}
    \newcommand\NOTE[1]{\leavevmode\ignorespaces\unskip}
\fi

\allowdisplaybreaks

\newtheorem{definition}{Definition}

\DeclareMathOperator*{\argmax}{\arg\max}
\DeclareMathOperator*{\argmin}{\arg\min}

\algnewcommand{\IIf}[1]{\State\algorithmicif\ #1\ \algorithmicthen}
\algnewcommand{\ElseIIf}[1]{\State\algorithmicelse\ }
\algnewcommand{\EndIIf}{\algorithmicend\ \algorithmicif\ }
\algnewcommand{\EndIIff}{\algorithmicend\ \algorithmicif\ }

\makeatletter
\newcommand{\manuallabel}[2]{\def\@currentlabel{#2}\label{#1}}
\renewcommand{\@IEEEsectpunct}{.\ \,}
\renewcommand\subsubsection{\@startsection{subsubsection}{3}{\z@}%
                                     {0.0ex plus 0.2ex minus 0.2ex}%
                                     {0ex}
                                     {\normalfont\normalsize\bfseries}}
\makeatother

\author{Gabriele~Castellano, %
        Flavio~Esposito
        and~Fulvio~Risso
\thanks{G. Castellano and F. Risso are with the Department of Control and Computer Engineering, Politecnico di Torino, Italy, \texttt{gabriele.castellano@polito.it}, \texttt{fulvio.risso@polito.it}.}%
\thanks{F. Esposito is with the Computer Science Department at Saint Louis University, USA, \texttt{flavio.esposito@slu.edu}.}}%

\author{ 
	\IEEEauthorblockN{
		Gabriele Castellano\IEEEauthorrefmark{2}\IEEEauthorrefmark{1}, Juan-José Nieto \IEEEauthorrefmark{3}\IEEEauthorrefmark{4}, Jordi Luque \IEEEauthorrefmark{3}, Ferrán Diego \IEEEauthorrefmark{3}, Carlos Segura \IEEEauthorrefmark{3}\\ 
        Diego Perino \IEEEauthorrefmark{3}, Flavio Esposito\IEEEauthorrefmark{2}, Fulvio Risso\IEEEauthorrefmark{1}, Aravindh Raman \IEEEauthorrefmark{3}
		\\
		\IEEEauthorblockA{
			\IEEEauthorrefmark{2}Computer Science, Saint Louis University, USA
		}
		\IEEEauthorblockA{
			\IEEEauthorrefmark{1}Computer and Control Engineering, Politecnico di Torino, Italy
		}
         \IEEEauthorblockA{
			\IEEEauthorrefmark{4}Universitat Politècnica de Catalunya, Spain
        }
        \IEEEauthorblockA{
			\IEEEauthorrefmark{3}Telefónica Research, Spain
        }
		Email: \IEEEauthorrefmark{2}\{gabriele.castellano\}@slu.edu, \IEEEauthorrefmark{3}\{jordi.luque\}@telefonica.com
	}
}

\IEEEoverridecommandlockouts
\begin{document}
%
\title{\LARGE{Scheduling Inference Workloads on Distributed Edge Clusters with Reinforcement Learning}}

\maketitle


\begin{abstract}
%
Many real-time applications (e.g., Augmented/Virtual Reality, cognitive assistance) rely on Deep Neural Networks (DNNs) to process inference tasks. Edge computing is considered a key infrastructure to deploy such applications, as moving computation close to the data sources enables us to meet stringent latency and throughput requirements. However, the constrained nature of edge networks poses several additional challenges to the management of inference workloads: edge clusters can not provide unlimited processing power to DNN models, and often a trade-off between network and processing time should be considered when it comes to end-to-end delay requirements. In this paper, we focus on the problem of scheduling inference queries on DNN models in edge networks at short timescales (i.e., few milliseconds). By means of simulations, we analyze several policies in the realistic network settings and workloads of a large ISP, highlighting the need for a dynamic scheduling policy that can adapt to network conditions and workloads. We therefore design \algname, a Reinforcement Learning based scheduling algorithm able to adapt its decisions according to the system conditions. Our results show that \algname\ effectively provides the best performance compared to static policies when scheduling over a distributed pool of edge resources.
\end{abstract}



%


\section{Introduction}
\label{sec:introduction}
In the last years, we have witnessed the growing popularity of applications leveraging Deep Neural Networks (DNNs), from Augmented/Virtual Reality (AR/VR) to cognitive assistance or video surveillance. The DNN model training process typically does not have strict latency constraints and it is performed \emph{offline} in well-provisioned centralized data-centers or in a distributed fashion via, e.g., federated learning~\cite{konecny2016federated-learning}.
Differently, the DNN inference task is usually performed \emph{online} with constraints in terms of accuracy, throughput, and latency, which may significantly differ across applications.
For instance, services like cognitive assistance require high accuracy but may tolerate few hundreds of milliseconds latency, while others, like self-driving cars, have more stringent latency needs (i.e., tens of milliseconds).

{\let\thefootnote\relax\footnote{{* Gabriele Castellano and Juan-José Nieto contributed equally to this work during his internship at the Telefónica Research team, in Spring 2020.}}}
Providing an inference service requires to address several challenges to meet this diverse set of application constraints, e.g., the selection of the appropriate variant of the model to be used (programming framework, compiler optimization, batching size, etc.), the processing unit to leverage for the inference (e.g., GPU, CPU, TPU), and the nodes and resources (e.g., memory, computing) to be allocated to every application. This requires management at different timescale. On a short timescale (i.e, milliseconds), a \emph{scheduler} is in charge of selecting the appropriate computing instance for every new incoming request to meet its application requirements. This includes not only the selection of the computation node but also the appropriate model variant and computation technology. On a longer timescale (i.e., seconds, minutes), an \emph{orchestrator} selects the proper model variants to deploy, optimizes their placement across the nodes, and allocates the appropriate resources to them. 
Recent work~\cite{kannan2019grandslam,mao2019learning,crankshaw2017clipper,romero2019infaas} focused on data centers and proposed DNN inference workload management for such environments. Further, commercial solutions have been deployed in recent years~\cite{olston2017tensorflow,chappell2015introducing,enginegoogle} by major cloud providers.

Edge computing is considered a key enabler to deploy DNN-based applications with stringent delay or bandwidth requirements, as it moves computation capabilities closer to end-users with respect to centralized cloud platforms. This is especially the case for users connected via mobile access (e.g. 5G). However, realizing DNN inference at the edge poses several additional challenges. Edge infrastructures are indeed complex networks composed of several layers with heterogeneous limited resources and different latencies to end users~\cite{mach2017mobile}.
Due to the less availability of resources at edge, multiple inference models of different capacities should be considered, and end-to-end delay requirements may lead to considering a trade-off between network delay and processing time. This differs from centralized cloud platforms, which usually feature large pools of uniform hardware available in a single location where DNN models can be scaled up almost indefinitely. 
For these reasons, the optimal selection of inference models while scheduling real-time requests at Edge is still a challenging task.
Recent work combined edge computing and deep learning~\cite{chen2019deep}, with a focus on 
scheduling requests to minimize end-to-end delay~\cite{ghosh2018distributed} or 
maximize accuracy~\cite{hung2018videoedge}.
However, none of the existing work analyzes inference workload optimization taking into account different application constraints in realistic edge network settings.

In this paper, we focus on the problem of \emph{scheduling} DNN inference requests taking into account not only accuracy (i.e., model selection) but also throughput and latency constraints under realistic edge deployment settings. First, we model our distributed edge inference system and provide a definition of the scheduling problem (Section~\ref{sec:design}), also proposing several baseline static scheduling policies both original and from literature. From evaluating static policies on a realistic network topology, we observe that a policy that always performs better does not exist, as different applications may benefit differently from each scheduling strategy.
Based on the insights derived by this analysis we propose \algname\footnote[1]{
In ancient Egyptian mythology, \algnamelower\ was a major goddess said to have power over fate itself. 
}
(\algfullname), an adaptive scheduling algorithm based on Reinforcement Learning (Section~\ref{sec:scheduling}), which dynamically follows system conditions and apps requirements optimizing its decisions accordingly.
We evaluate \algname\ simulating three topologies based on the realistic network of a large ISP and using a pool of reference edge applications (Section~\ref{sec:validation}). Our findings show that, while some static policies are well suited to optimize workloads on cloud-based topologies, \algname\ improves performance over any static policy when resources are distributed across the edge network, effectively increasing the percentage of successfully handled queries.

\section{Related Work}
\label{sec:related-work}

The provisioning of on-demand inference services has been investigated in several recent works. 

\noindent \textbf{Inference scheduling in data centers}. Most of the existing solutions address the common scenario where inference queries have to be scheduled over the resources of a Data Center. Some of the main production systems are Tensorflow Serving~\cite{olston2017tensorflow}, Azure ML~\cite{chappell2015introducing}, and Cloud ML~\cite{enginegoogle}. Most scientific works focused on proposing algorithms and strategies to improve the performance and ease of use of such cloud inference systems.~\cite{kannan2019grandslam} and~\cite{mao2019learning} address the problem of scheduling Directed Acyclic Graph (DAGs) tasks with the objective of improving the throughput; GrandSLAm~\cite{kannan2019grandslam} relies on a prediction model that estimates job duration, while~\cite{mao2019learning} proposes an efficient RL approach to select the number of servers to allocate for a given job. Being oriented to a Cloud infrastructure, none of them takes into account network latency between the servers and their heterogeneity.  In~\cite{soifer2019deep} a Model Master manages the dynamic allocation of DNN models across the servers of a heterogeneous data center based on Azure ML, and proposes a protocol among servers to forward queries to the correct destination. Clipper~\cite{crankshaw2017clipper} provides a generalization of TensorFlow Serving~\cite{olston2017tensorflow} to enable the usage of different frameworks.
One of the most complete solutions is provided by INFaaS~\cite{romero2019infaas}, which focuses on ease of use, providing transparent scheduling of incoming queries over available model variants, and autoscaling of deployed models based on load thresholds. 
However, all the previous works address the scheduling problem only from the boundaries of a data center, considering neither \textit{(i)} network latency, thus becoming no suitable in scenarios with real-time constraints, nor \textit{(ii)} resource constrained clusters, thus failing to address situations where workers cannot be indefinitely scaled up/out.

\noindent\textbf{Inference offloading}. Another related set of works concerns offloading, with a focus on the end-devices. While offloading has been widely studied in the literature~\cite{cuervo2010maui, ra2011odessa}, the specific use case of DNN workload introduces additional degrees of freedom (e.g., model variant selection and configuration) that can be exploited for improving optimization over the mere selection of the task placement. Some recent works~\cite{han2016mcdnn, ran2018deepdecision, hu2019linkshare} provides intelligent offloading techniques for DNN tasks. DeepDecision~\cite{ran2018deepdecision} addresses the problem in the particular case of a single device running a single application; queries are scheduled among a series of local small models providing different performance/requirements trade-off, and one remote model, which provides the best performance.
On the other hand, LinkShare~\cite{hu2019linkshare} focuses on the orthogonal problem of ordering the offloaded requests from multiple apps on the same device, with the main constraint of network bandwidth.
MCDNN~\cite{han2016mcdnn} proposes a scheduler to handle queries from multiple applications on the same device, deciding~\textit{(i)} the model variant to be used and~\textit{(ii)} whether to offload the inference task or not, seeking average accuracy maximization. Such decisions are taken considering constraints such as latency requirements, device energy, cloud monetary budget.

\noindent \textbf{Inference and edge computing}. Fewer and more recent are the trends that combine DNN with edge computing~\cite{chen2019deep}, with the aim of overcoming scalability and latency limitations of cloud computing. The use of edge computing brings additional challenges deriving from the high resource requirements of DNN based tasks on less powerful edge compute resources. Despite some issues have been addressed in recent works~\cite{mortazavi2017cloudpath, khare2019linearize, hung2018videoedge, ghosh2018distributed}, edge-oriented solutions for inference systems are still largely embryonic compared to data center solutions, with many open challenges.
CloudPath \cite{mortazavi2017cloudpath} focuses on the problem of data distribution on a hierarchical continuum of computing resources between edge and cloud.
In~\cite{khare2019linearize}, authors propose an approach to schedule DAGs across multiple edge servers, seeking minimization of end-to-end latency. However, the proposed algorithm assumes the possibility to indefinitely allocate new edge servers when needed, with no geographical restrictions, thus not addressing the problem of constrained resources at the edge.
Other works~\cite{hung2018videoedge, ghosh2018distributed} study the problem of processing data streams from scattered devices, exploiting the geographically distributed edge/cloud clusters. In particular, VideoEdge~\cite{hung2018videoedge} assumes a deployment of cameras generating a known set of video streams, on which various DNN tasks should be performed. The proposed approach decides globally the cluster where each stream should be processed, as well as the model variant to employ and its configuration, considering computation and network bandwidth as constraints and seeking accuracy maximization. However, neither processing nor network latencies are taken as constraints, thus making this approach not suitable for interactive or critical scenarios (e.g., virtual reality, autonomous driving, and more). 
A similar use case is analyzed in~\cite{ghosh2018distributed}, which focuses on minimizing the end-to-end latency processing data flowing from the edge to the cloud. However, it only considers the problem of task allocation, missing the possibility to optimize properly selecting model variants and their configurations.

To the best of our knowledge, none of the existing works on inference serving systems addresses the problem simultaneously considering \textit{(i)} end-to-end latency, accuracy, and throughput constraints, \textit{(ii)} edge-cloud computing and multi-cluster deployment, \textit{(iii)} real-time job dispatching, \textit{(iv)} optimization on model variant selection.
\section{Scheduling in edge-cloud infrastructure}
\label{sec:design}

In this section, we formally define the problem of scheduling inference tasks on a distributed edge-cloud infrastructure. Additionally, we describe a set of static scheduling policies (both original and from literature), that we then use in Section~\ref{sec:scheduling} as a baseline for our dynamic scheduling approach.

\subsection{System modeling}
\label{subsec:design-settings}

\noindent\textbf{Applications and data-streaming sources.} 
We consider a set of sources (e.g., end users, IoT devices, vehicles) running a variety of applications (e.g., virtual reality, autonomous driving) each relying on one or more DNN inference tasks. 
Every application generates \textit{queries} to be processed, i.e., each query represents the request to perform a specific inference task $j \in J$ (e.g., object detection, speech recognition) on a given input (e.g., a video frame), where $J$ is the set of inference tasks supported by the system. Since applications often require more than one query to be processed, we treat sequential queries as streams (e.g., all the frames captured by an AR headset). Therefore, each query $q$ belongs to a stream $i\in I$, being $I$ the entire set of streams currently served by the system. Every query of a stream has a set of requirements such as a maximum end-to-end delay $D^i$, and a minimum required accuracy $A^i$. Additionally, every stream $i$ has a \textit{data rate} $\rho_i$, that is the number of queries submitted each second (e.g., frame rate), and every query of stream $i$ has an input of  \textit{size} $\zeta_i$ (e.g., frame size). Note that all queries of a stream are for the same task $j \in J$ with the same requirements. 

\noindent
\textbf{DNN Models and Variants.} 
Every inference task $j$ can be served using a \emph{Deep Neural Network model} $m$ among the set of $M^j$ models that are trained for task $j$. Therefore, the system provides a total of $N_m=\sum_{j \in J} |M^j|$ DNN models.
Take object detection as an example application. A model $m$ represents a particular Neural Network architecture with pre-trained weights (e.g., yolo-v3, ssd-mobilenet-v1), and features a given accuracy $A_m$ (mean average precision - mAP). A model $m$ can be deployed and run through different setups and underlying hardware (e.g., SSD Mobilenet v1 on \textit{(i)} Tensorflow-GPU with batch size 8, or on \textit{(ii)} Opencv-CPU batch size 1 and 2 replicas, and more), thus obtaining a set $V^m$ of different \textit{model variants}. A model variant $v$ features a given \textit{processing delay} $D_v$, throughput \textit{capacity} $C_v$ (i.e., the maximum number of queries it can process per second), and \textit{resource usage} $\boldsymbol{r}_v \in \mathbb{R}_+^k$ (e.g., in terms of CPU, system memory and GPU memory). 
Note that the processing delay may vary based on the size $\zeta_i\in\mathbb{R}_+$ of the input data, thus it is a function $D_v \colon \mathbb{R}_+ \to \mathbb{R}_+$; with $D_v$ we refer to the time needed to process the maximum input size supported by the model (analogous considerations hold for the capacity $C_v$).

\noindent
\textbf{Network topology and computing clusters.} 
We consider a geographically distributed cloud-edge infrastructure composed by $N_\nu$ computing \textit{clusters} (e.g., a centralized data center, a telco regional cloud, an eNodeB) typically organized in a hierarchical topology. Each cluster potentially provides different resources. We denote $\boldsymbol{c}_n \in\mathbb{R}_+^k$ the overall capacity of cluster $n$, with $c_{nk}$ representing the amount of resource $k \in \mathbb{N}$ available on cluster $n$. 
Examples of resources include CPU, system memory and GPU memory.

Model variants are deployed at different computing clusters consuming a different amount of resources.  On a long timescale (i.e., seconds, minutes), an \emph{orchestrator} selects the appropriate set of model variants to deploy, optimizes their placement across the clusters, and allocates the appropriate resources. 
Finally, stream sources are connected to a small cluster at the lower layer of the hierarchy. This can be either the antenna/eNodeB in case of cellular communication or the home gateway in the fixed access case. Queries need to be \emph{scheduled} for processing across model variants available at different computing clusters to meet application requirements on a short timescale (i.e., tens of milliseconds).
In the following, we provide a definition of the scheduling problem we tackle in this paper.




\begin{figure}[t]
	\centering
	\includegraphics[clip= true, width=\columnwidth, bb=0 0 120 120, trim= 0.0in 3.3in 3.3in 0.0in]{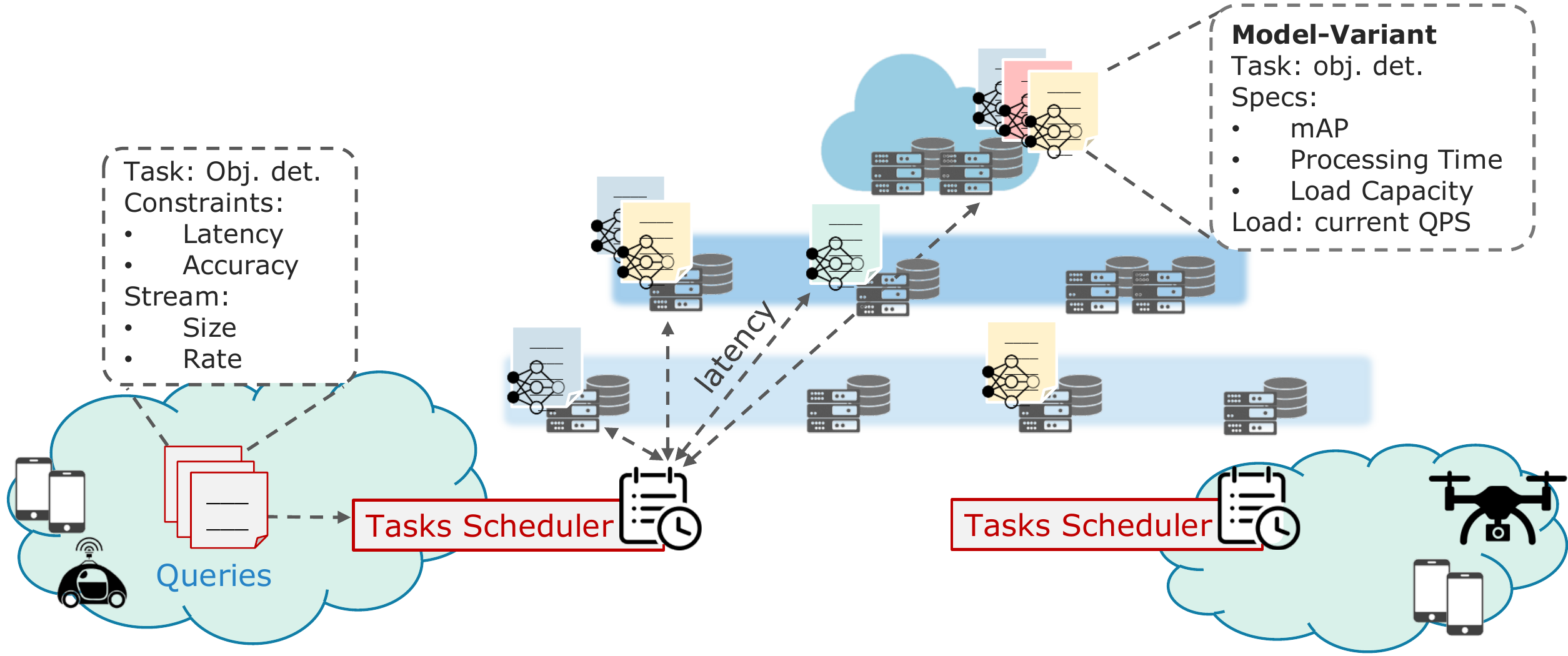}
    \caption{The scheduler dispatches streams of queries on available model variants based on their constraints and geographical position of clusters.}
	\label{fig:problem}
\end{figure}

%

\subsection{Scheduling problem definition}
\label{subsec:design-scheduling}
We assume a scheduler is located at the nearest compute cluster available to existing stream sources, i.e., antenna/eNodeB or the home gateway/central office in the fixed access case. It follows every stream source is served by a \textit{scheduler} $s$ among $N_s$ different ones (one per each lower layer cluster). 
Each scheduler $s$ has a given average network delay $d_n^s$ towards  each cluster $n$; we also model the associated delay deviation as ${\sigma}_n^s$. Note that an additional access delay from the stream source to the scheduler has to be taken into account (e.g, the radio latency between a device and the nearest 5G antenna). We denote $\delta_i$ the additional access delay that affects stream $i$.
Every scheduler is aware of each model variant $v$ currently available on each cluster $n$, each with its current load $L_{vn}(t)$ (measured in terms of incoming queries per second%
\footnote{Each stream $i$ contributes to the total load as a fraction $\eta_{v}^i$ of its data rate~$\rho_i$ (named \textit{fractional load)}, based on the stream data size $\zeta_i$.}).
Based on the current conditions of the available model variants, for every stream $i$ it serves, a scheduler $s$ decides which model variant $v$ on which cluster $n$ should be used to process stream $i$.

When scheduling a stream $i$ to the proper model variant/cluster, the scheduler takes into account application requirements. Specifically, it considers the stream data size $\zeta_i$, its data rate $\rho_i$, its bit rate $b_i$, the maximum tolerated end-to-end delay $D^i$ and the minimum required accuracy $A^i$, satisfying the following constraints:

\noindent
\textit{(i)} the selected model variant $v$ is a valid implementation of task $j$ required by $i$,
\begin{equation}
    \label{eq:compatible-variant}
    v \in V^m \land m \in M^j;
\end{equation}

\noindent
\textit{(ii)} the load capacity of the chosen model variant is not exceeded,
\begin{equation}
    \label{eq:load-constraint}
    L_{vn}(t) + \eta_{v}^i\rho_i \leq C_v,
\end{equation}
being $\eta_{v}^i$ the fractional load of stream $i$ for model variant $v$;

\noindent
\textit{(iii)} the sum of expected network delay and processing time does not exceed the maximum tolerated delay,
\begin{equation}
    \label{eq:latency-constraint}
    2(\delta_i + d_n^s + 2{\sigma}_n^s) + b_i \zeta_i + D_v(\zeta_i) \leq D^i,
\end{equation}
where the first addendum is the round-trip propagation time, the second is the transmission delay for one query and the third is the time needed to process the query;

\noindent
\textit{(iv)} the selected model provides an adequate accuracy 
\begin{equation}
    \label{eq:accuracy-constraint}
    A_m \geq A^i.
\end{equation}

A graphical representation of the scheduling problem is depicted in Figure~\ref{fig:problem}, while a scheduling policy can be formally defined as follows.

\begin{definition}
	\label{def:scheduling-policy}
	(scheduling policy). Let us consider a stream $i$ to be processed through a task $j$ on an edge-cloud infrastructure that features a set of $V^m$ compatible model variants over $N_\nu$ clusters ($|N| = N_\nu$). A scheduling policy is any function
	\begin{equation}
		\label{eq:policy}
		\begin{gathered}
            \mathcal{\beta} \colon I \to V^m, N
		\end{gathered}
	\end{equation}
	that binds stream $i$ to a feasible model variant $v \in V^m$ deployed on cluster $n \in N$, so that constraints at Equations~(\ref{eq:compatible-variant}), (\ref{eq:load-constraint}), (\ref{eq:latency-constraint}), and (\ref{eq:accuracy-constraint}) are satisfied.
\end{definition}

Note that, as requests are handled in real-time, scheduler decisions should be taken in an amount of time that is negligible compared to the stream latency requirements.

\noindent\textbf{Scheduling performance metrics and objectives.}
Based on the scheduling decisions, in a given time instant $t$ the stream $i$ will feature a \textit{reject ratio} $q_i^R(t)\in[0,1]$, i.e., the fraction of queries from stream $i$ that have not been processed by the system because of resource unavailability, and a \textit{failure ratio} $q_i^F(t)\in[0,1]$, i.e. the fraction of queries that have been served violating one or more application requirements (i.e., delivered out of maximum tolerated delay). 

The goal of the scheduler is typically to maximize, over time, the fraction of queries that are served successfully, i.e., to minimize the sum of reject ratio and failure ratio.
%

\subsection{Static scheduling policies}
\label{subsec:policies}

Several policies have been proposed for static scheduling of inference tasks on edge clusters~\cite{cicconetti2018architectural, mach2017mobile}. 
In this work we consider the following ones (both original and from literature):
\dpnote{In the policies below set a reference to the paper they have been derived from. Some of them are baseline, some of them state of the art.}

\textit{1) closest:} bind stream $i$ to any feasible model variant $v^*$ located on the cluster $n^*$ that features the lower network latency to serving scheduler $s$, i.e., $n^* = \argmin_{n \in N}{(d_n^s + 2{\sigma}_n^s)}$. This policy may lead to the early saturation of smaller clusters at the very edge, as they are always preferred~\cite{jia2015optimal}.

\textit{2) load balancing:} bind the input stream to model variant $v^*$ on cluster $n^*$ such that $(v^*, n^*) = \argmin_{v, n \in V^m \times N}{L_{v n}(t)}$. This policy can bring huge performance gains compared to \textit{closest}~\cite{jia2015optimal}; however, it may lead to unfair allocation when latency-sensitive applications are in the minority.

\textit{3) farthest:} bind stream $i$ to any feasible model variant $v^*$ located on the cluster $n^*$ with the highest (still feasible) network latency, i.e. $n^* = \arg\max_{v \in N}{(d_n^s + 2{\sigma}_n^s)}$. As opposed to \textit{closest}, this policy preserves smaller clusters at the very edge for those apps that really need them~\cite{long2015green}; however, it is highly affected by the unreliability of network delay for long distance communications.

\textit{4) cheaper:} bind stream $i$ to model variant $v^*$ on cluster $n^*$ such that the expected end-to-end delay (round-trip and processing time) is maximized, i.e., $(v^*, n^*) = \argmin_{v, n \in V^m \times N}{(2(d_n^s + 2{\sigma}_n^s) + D_v(\zeta_i))}$. We designed this policy as an improvement over \textit{farthest}, as it additionally tries to preserve the most performing model variants.

\textit{5) random-proportional latency:} bind stream $i$ to model variant $v$ on cluster $n$ with probability $1/(2(d_n^s + 2{\sigma}_n^s) + D_v(\zeta_i))$. This guarantees that, on a large enough number of streams, bindings are proportionate to end-to-end delays~\cite{cicconetti2018architectural}.

\textit{6) random-proportional load:} bind stream $i$ to model variant $v$ on cluster $n$ with probability $C_v/L_{v n}(t)$. This guarantees that, on a large enough number of streams, bindings are proportional to the capacity of each model variant.

\textit{7) least impedance:} 
bind stream $i$ to model variant $v^*$ on cluster $n^*$ such that end-to-end latency to $s$ is minimized, i.e., $(v^*, n^*) = \argmin_{v,n \in V^m \times N}{(2(d_n^s + 2{\sigma}_n^s) + D_v(\zeta_i))}$~\cite{cicconetti2018architectural}. This greedy policy leads to the best performance when the overall load is low, but may suffer from a high rejection rate once the closest and fastest model variants are saturated.

Our experiments (Section~\ref{sec:validation}) show that, for a heterogeneous pool of applications, a policy that always performs better than the others does not exists: different applications may benefit differently from each scheduling strategy, and also the physical topology and the particular streams arrivals can be determinant.
Based on these findings, in the next section we propose \algname, an algorithm for \algfullname\ that leverages Reinforcement Learning to optimize its decisions dynamically based on the current system conditions. 

\section{\algname\ Scheduling Algorithm}
\label{sec:scheduling}



Our adaptive scheduling approach aims to learn the optimal policy depending on current system conditions, e.g, current applications, network topology, and stream arrivals that vary over time. Due to the lack of labeled data, the optimal policy learning is formulated as a Reinforcement Learning (RL) problem; hence, an intelligent agent tries to learn the optimal policy selection strategy according to the observed state of the environment. This is accomplished by an RL policy that estimates a probability distribution
of each possible action (policy selection) that cumulatively maximizes a reward (typically  maximizing the fraction of queries that are served successfully), as shown in Figure 2. 

Let us consider a learner and decision-maker called the \textit{agent}, and an \textit{environment} that is the external world that the agent interacts with at discrete time steps $t$.
Given $S_t \in S$, where $S$ is the set of possible \textit{states} of the environment, the agent can select an \textit{action} $A_t \in A(S_t)$,
standing for the set of available actions in state $S_t$. The agent receives an observation of the environment $S_t$ at time $t$ and, one step later, a numerical \textit{reward},
$r_{t+1} \in R \subset \mathbb{R}$, and it jointly determines the action $A_t$ to perform, which, in part, yields to next state $S_{t+1}$.


\begin{definition}
	\label{def:RL-policy}
	(stochastic reinforcement learning policy). An RL policy $\pi_\phi$, where $\phi \in \mathbb{R}^d$ denotes policy parameters, is any function or algorithm that determines and maps the next action to take by an agent. A stochastic RL policy, additionally, estimates a probability distribution over actions that an agent can take at a given state: 
	%
		\begin{align}
		\label{eq:RL-policy}
		   &\pi_\phi \colon \: A \, \text{x} \, S \to [0,1], \\
           &\pi_\phi(a|s) \overset{\mathrm{def}}{=\joinrel=} \ensuremath{\mathbb{P}} (\text{take action } a |\text{given state } s).\nonumber
		\end{align}
\end{definition}

\begin{figure}[t]
        \includegraphics[clip= true, width= \columnwidth, trim={0.2cm 15.7cm 16.1cm 0.1cm}]{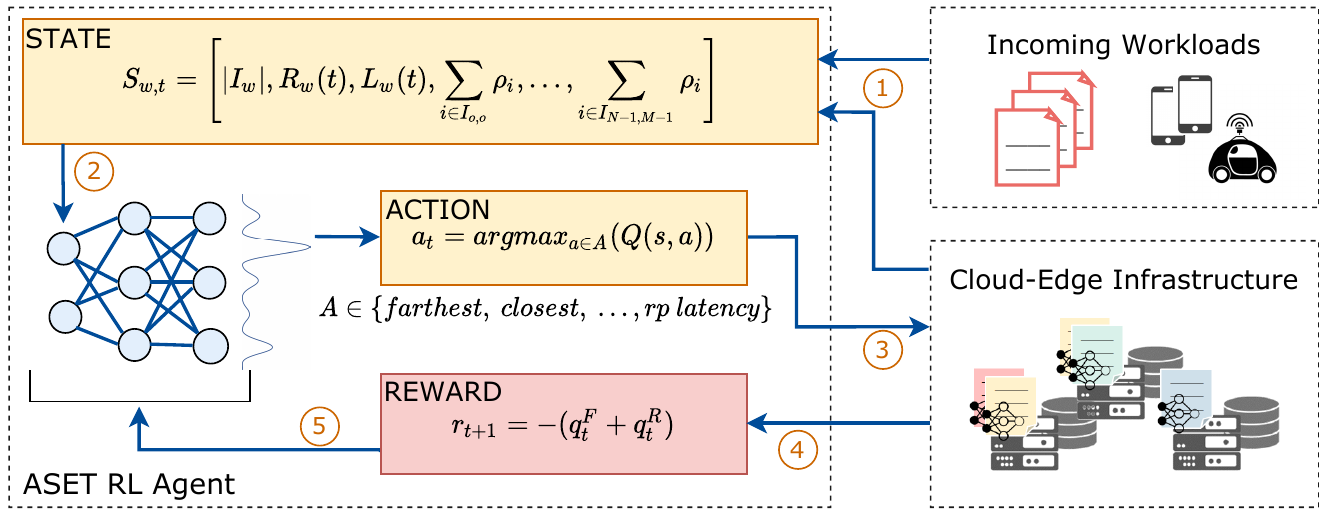}
    \caption{Algorithm overview. State $S_t$, sampled from the environment, is forwarded through the agent DNN, which outputs action $A_t$; performing $A_t$ on the environment contributes to reward $r_{t+1}$ obtained at the next step.}
    \label{fig:overview}
\end{figure}

Overall, the goal of the proposed adaptive scheduling is to learn an optimal sequence of static network scheduling policies that maximizes the percentage of successfully dispatched streams. At a $T$ seconds rate, the RL-based scheduler samples the environment by collecting a variety of observations from the edge-cloud infrastructure, e.g., responses and loads, building up the current state $\mathit{S_t}$ of the environment. Then, the agent evaluates a discrete set $A$ of actions and chooses an action $A_t \in A$, where $A$ stands in this work for the set of available network scheduling policies $\mathcal{\beta}$. Note that the set of actions does not depend on the state itself, thus the sets $A(S_t) = A$ are the same (Section~\ref{subsec:policies}). Therefore, every time that the agent takes an action $\mathit{A_t}$, the state of the environment $\mathit{S_t}$ is observed and a reward score $\mathit{r_
{t+1}}$ is used as feedback information to improve the policy selection, see Figure~\ref{fig:overview}. In this work, these rewards are defined as a linear combination of the ratio of ``failed'' queries and the ratio of queries that have been ``rejected'' for lack of available resources (Section~\ref{subsec:training}).

\begin{figure}[t]
    \centering
    \begin{subfigure}[t]{0.49\columnwidth}
        \centering
        \includegraphics[clip= true, width=\linewidth, trim=0in 0in 0in 0in 4.2in]{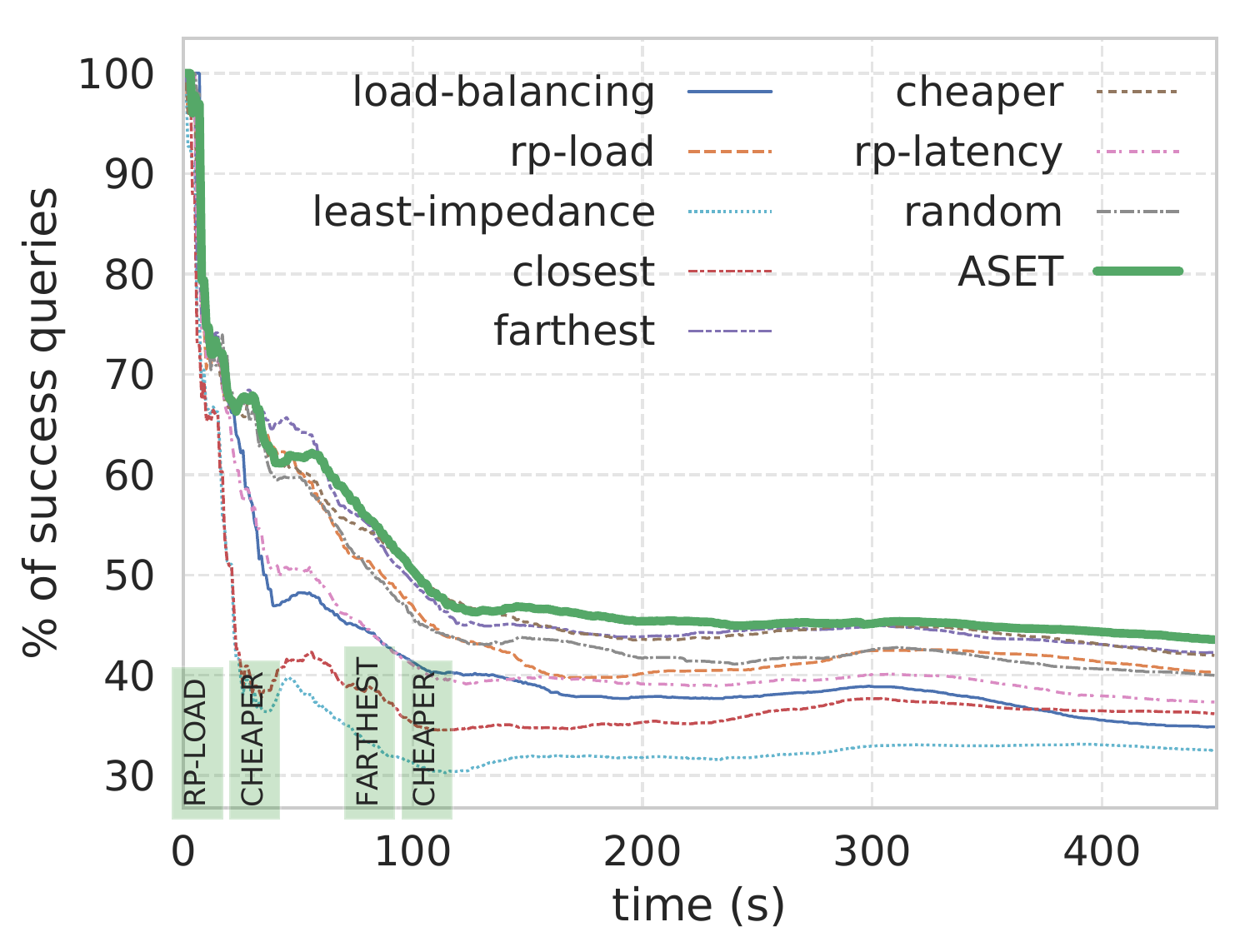}
        \vspace{-6mm}
        \caption{Cloud-based topology.}
        \label{subfig:policy-switching-co}
    \end{subfigure}%
    \begin{subfigure}[t]{0.49\columnwidth}
        \centering
        \includegraphics[clip= true, width=\linewidth, trim=0in 0.0in 0in 0in]{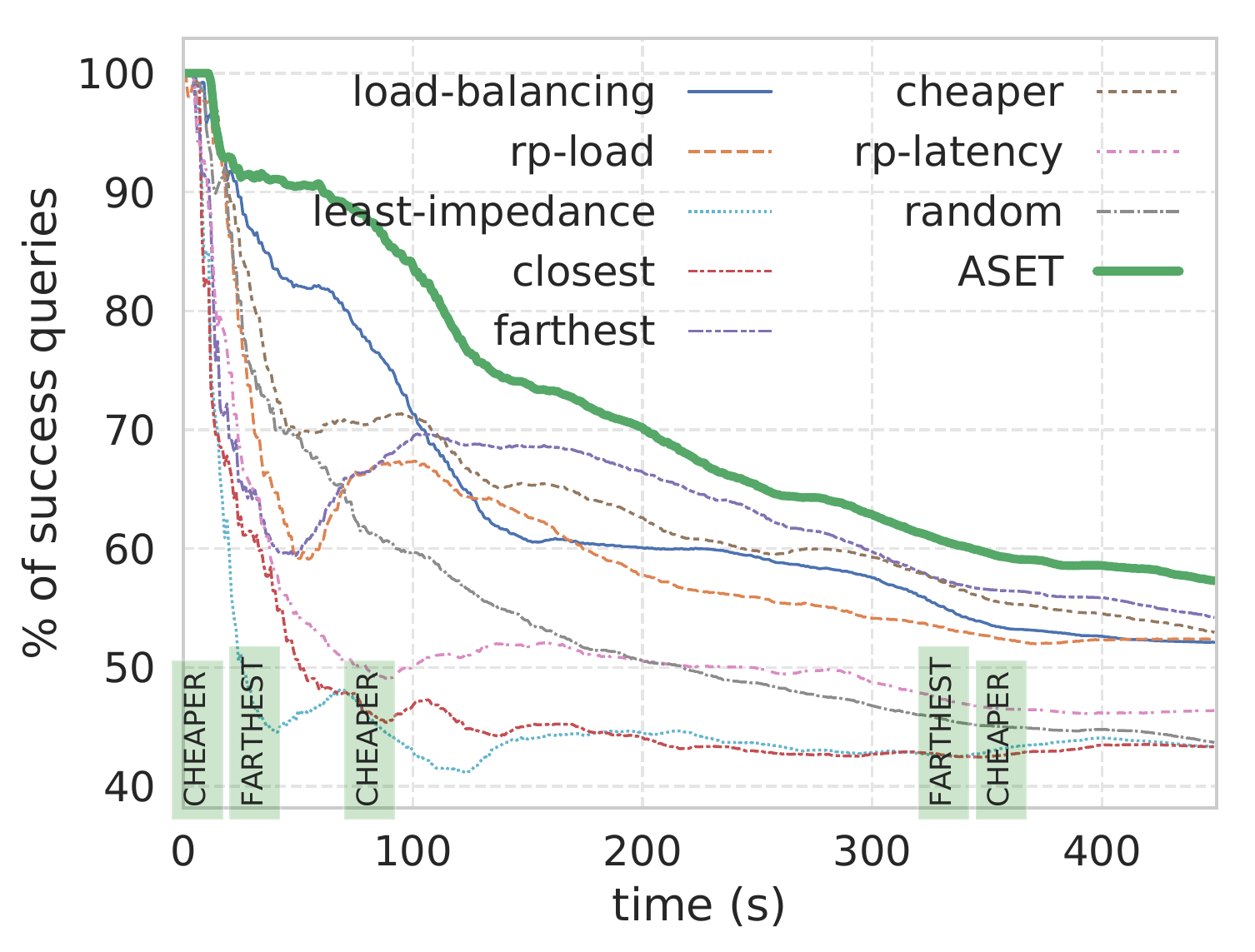}
        \vspace{-6mm}
        \caption{Edge-based topology.}
        \label{subfig:policy-switching-edge}
    \end{subfigure}
    \caption{
    The \algname\ RL agent infers the optimal policy sequence based on the system conditions, seeking an optimal binding between workloads and model variants that maximizes the percentage of success queries. Plots show two runs on a cloud-based topology and on an edge-based one (see Section~\ref{sec:validation}).
    }
    \label{fig:policy-switching}
\end{figure}

The particular policy $\mathcal{\beta}_t$, selected by the agent at time $t$, is used to dispatch all incoming streams during the subsequent time window $[t,t+T]$. Therefore, given the corresponding states sequence $\mathbf{S} = [{S}_0, {S}_T, {S}_{2T}, ... , {S}_{kT}]$ with $k \in \mathbb{N}$, the resulting overall scheduling policy $\mathcal{\beta}(\mathbf{S}) = [\mathcal{\beta}_0, \mathcal{\beta}_T, \mathcal{\beta}_{2T}, ... , \mathcal{\beta}_{kT}] $ dynamically maps, with the corresponding baseline policies $\mathcal{\beta}_t$, a stream $i$ to a model variant $v$ and its deployment on cluster $n$. From now, and for the sake of simplicity, we will refer as $\pi$ to the policy learned by the \algname\ agent (Definition~\ref{def:RL-policy}), which leads to a particular static policy sequence $\mathcal{\beta}(\mathbf{S})$. It corresponds to any function employed to estimate the optimal sequence of actions that the agent should perform at each time window $[t,t+T]$ and given a state $S_t$, $\mathcal{\beta}(\mathbf{S}) = [\mathit{A}_0, \mathit{A}_T, \mathit{A}_{2T}, ... , \mathit{A}_{kT}]$.
%
The intuition of this behavior is provided in Figure~\ref{fig:policy-switching}.
Note that each of the static scheduling policies from Section~\ref{subsec:policies} corresponds to a deterministic agent that always returns the same action $\mathit{A_t}$ independently of the system state; whereas the policy $\pi$ learned by the \algname\ agent can be seen as a meta-policy (or as a policy of baseline scheduling strategies) that also satisfies the constraints from Equations~(\ref{eq:compatible-variant}), (\ref{eq:load-constraint}), (\ref{eq:latency-constraint}), and (\ref{eq:accuracy-constraint}).\looseness=-1

\vspace{-0.1cm}
\subsection{Deep Q-Learning policy optimization}
\label{subsec:q-learning}

Our RL agent has to cope with a discrete set of actions, with $A \subset \mathbb{N}$. This is often modeled in literature as a stochastic process with no memory, which is a Markov Decision Process~\cite{bellman1957markovian} (MDP). In this work, our MDP defined by tuples $(S,A,\mathcal{T},\mathcal{R},\gamma$) represents states comprised of partial observations from the system. Nonetheless, the model parameters of such MDP are unknown, i.e., the transition probabilities ${\mathcal{T}(s'|a,s)}$ and the rewards ${\mathcal{R}(s'|a,s)}$ of taking the action $A_t=a$ and moving from state ${S_t}=s$ to state ${S_{t+1}}=s'$. Note that the \algname\ agent should experience each transition among states at least once, or even multiple times to get a reliable estimation of both transition and cumulative rewards. 


At each step $t = kT$, with $k \in \mathbb{N}$, the RL agent can choose one of several possible scheduling policy-actions, $\mathcal{\beta}_t \equiv \mathit{A_t}$. The transition probability $\mathit{\mathcal{T}(s'|a,s)}$ depends in part on the chosen action, and, additionally, 
\gc{from some positive or negative reward that may be returned by every state transition, named \textit{return} of actions.}
Overall, our objective is to find a strategy, i.e., a policy $\mathcal{\pi}$ mapping to a sequence $\mathcal{\beta}(\mathbf{S})$, that maximizes the expected return $G(t)$ of rewards over time. Thus, $G(t)$ is defined in terms of the cumulative weighted rewards along with states and given the corresponding optimal sequence of actions to take in the future:
\begin{equation}
\label{eq:return}
    G(t) = \sum_{\tau=0}^{H} \gamma^\tau r_\tau \qquad \gamma \in [0,1], 
\end{equation}
where $r_\tau = {\mathcal{R}(s'|a,s)}$ is the reward at time step $\tau$ due to corresponding state transition $(s,s')$, $\gamma$ is a weighting factor that reduces the contribution of long-term rewards, usually known as the discount factor, and time $H$ is the last time step within a training episode (see Section~\ref{subsec:training} for further details).
Therefore, the RL agent's target policy is
\begin{equation}
\label{eq:optimal-policy}
    \pi^{*}(a|s) = \argmax_{\pi_\phi}\mathbb{E}_{t^{*} \sim \pi_\phi}\left\lbrace G(t)\right\rbrace,
\end{equation}
which translates the scheduler state into a distribution over actions, see Definition~\ref{def:RL-policy}. Note that the expectation is computed over the distribution of trajectories 
$t^{*} = (s_0,a_0,s_1, ...)$.

In Q-Learning, the optimal pair values $(s,a)$, i.e., those yielding to the sequence of optimal actions, are generally called Quality-Values (Q-Values) and noted as $Q^*(s,a)$~\cite{levine2020}. They correspond to the sum of weighted rewards that the RL agent can expect on average after performing action $a$ on state $s$. It is also known as the \textit{expected return of actions},
\begin{equation}
    \label{eq:qvalue}
    Q(s,a) = \mathbb{{E}}_{t^* \sim \pi_\phi}\left\lbrace G_t|S_t = s,A_t=a\right\rbrace.
\end{equation}
Bellman~\cite{bellman1957markovian} showed that if an agent's trajectory follows the highest Q-Values, then its policy is optimal and leads to the highest $G(t)$ as well. Bellman also reported that an accurate estimate of Q-Values can be found recursively by using the \textit{Bellman Optimality Equation}, also known as the Value Iteration algorithm. In fact, Q-Learning is an adaptation of Bellman's value iteration algorithm, where a policy is implicitly, or off-line, learned by following trajectories yielding to the highest Q-Values~\cite{levine2020}. It is usually computed by dynamic programming and assumes that the optimal value of state $S_t = s$ \luq{juanjo, check time indexes for s and s' and within equation 10. Here maybe mention TD trick to link with training section} is equal to the reward it will get on average, after taking one optimal action $a$ and adding the expected optimal value of all possible next states along the future path of decisions, that is $ Q(s,a)  = \mathbb{E_\pi}\left\lbrace r+\gamma\max_{a'}Q(s',a')|s,a\right\rbrace $.
Equation~(\ref{eq:qvalue}) turns out into the following iteration algorithm, which converges to the optimal $Q^*(s,a)$,
%
\begin{equation}
    \label{eq:qvalue-ite}
    Q_{k+1}(s,a) \gets \sum_{s'} \mathcal{T}(s,a,s') \big[ r + \gamma \max_{a'}Q_k(s',a') \big],
\end{equation}
for all $s' \in S$, $a' \in A$ and $k \in \mathbb{N}$ as iteration step. For simplicity, \jj{we set the transition probability matrix $\mathcal{T}$ to all elements equal to $\mathit{1}$,}\jjdelete{we have initially set the transition probability matrix $\mathcal{T}$ to all elements equal to $\mathit{1}$,} allowing initial transitions among all seen states.
Once Q-Values are estimated, the optimal policy $\pi^{*}$ for the RL agent corresponds to chose the action that has the highest Q-Values: $\pi^{*}(a|s) = \argmax_{\pi} Q_{\pi}(s,a)$,
%
%
for all $s \in S$ and $a \in A \equiv \beta$ static policies in Section~\ref{subsec:policies}.

However, previous algorithm does not scale for large MDPs with a large number of states. A solution is to approximate the optimal $Q^*(s,a)$ using a Deep Neural Network, named Deep Q-Network (DQN)~\cite{dqn2013}, to get an estimate $Q(s,a;\phi) \approx Q^*(s,a)$, where $\phi$ \luq{better change greek letter? what u think guys?}\gabriele{I am fine with $\theta$ for representing the DNN}\carlos{me too, theta is usually used for the model parameters} stands for the parameters of the DQN model, see line 16 in Algorithm~\ref{alg:ASET}. The using of a DQN for approximate Q-Learning is known as Deep Q-Learning.

\vspace{-0.1cm}
\subsection{State Encoding}
\label{subsec:state}

We model the state in a continuous fashion, representing the environment in a given time $t$ as a set of some particular features sampled from the system and averaged along a time window of size $T$. Features are evaluated \carlos{computed, generated} separately for each available worker $w \in W$,%
\footnote{A worker is a model variant instance $v$ running on a particular cluster $n$, therefore we can assume index $w = n \cdot v + v$.} 
and are as follows:
\textit{(i)} the number $|I_{w}|$ of streams currently served by worker $w$, being $I_{w} = \{i \in I |~\beta(i) = (v, n)\}$;
\textit{(ii)} the current throughput $R_{w}(t)$ of worker $w$, in terms of responses delivered at the time instant $t$; 
\textit{(iii)} the current load $L_{w}(t)$, measured in terms queries per second normalized on input size (as defined in Section~\ref{subsec:design-scheduling});
\textit{(iv)} number of incoming instant queries grouped by stream characteristics, e.g., queries of all streams that require end-to-end delay within a given range $[\delta^1, \delta^2[$ and features a data rate in the interval $[\rho^4, +\infty[$, i.e., $\sum_{i \in I_{1,4}}\rho_i$, where $I_{1,4}=\{i \in I |~D^i\in [\delta^1, \delta^2[ \land \rho_i\in [\rho^4, +\infty[\}$.
In particular, we consider a partition $0 = \delta^0 < \delta^1 < \delta^2 < ... < \delta^{N_\delta-1}$ of $\mathbb{R}_+$ with $N_\delta$ delay intervals, and a second partition $0 = \rho^0 < \rho^1 < \rho^2 < ... < \rho^{N_\rho-1}$ of $\mathbb{R}_+$ with $N_\rho$ input-rate intervals, evaluating $N_\delta \cdot N_\rho$ different sum of instant queries, that is one feature for each combination of the two partitioning sets. The features defined so far constitute a vector as,
\begin{equation}
    \label{eq:feature-vector}
    \begin{gathered}
    \boldsymbol{S}_{w, t} =
    \Bigg[|I_{w}|,
    R_{w}(t),
    L_{w}(t),
    \sum_{i \in I_{0, 0}}\rho_i,
    \dots, \smashoperator[l]{\sum_{i \in I_{N_\delta-1, N_\rho-1}}}\rho_i\Bigg]
    \end{gathered}
\end{equation}
where $\boldsymbol{S}_{w, t} \in \mathbb{R}_+^{3+N_\delta \cdot N_\rho}$.
Therefore, the complete state $\boldsymbol{S}$ is modeled as a three-dimensional vector in $\mathbb{R}_+^{(3 + N_\delta \cdot N_\rho) \times |W| \times T}$, that is, each feature in (\ref{eq:feature-vector}) is first evaluated for each available worker (each model variant on each node), and then for each time instant within the considered time window. For instance, vector $\boldsymbol{S}_{w}$ stores, for worker $w$, every features in (\ref{eq:feature-vector}) evaluated at every time instant $t-T+1$, $t-T+2$, \dots, $t$ within time window $[t-T+1,t]$.
From now, we refer to the state vector encoding as simply $s$ or $s_t$ for a generally speaking state or a state referred to a time window, respectively.

\vspace{-0.1cm}
\subsection{Training}
\label{subsec:training}

The proposed RL scheduling agent is trained over a series of episodes that resemble various scenarios. Each episode corresponds to a different workload execution with given parameters, e.g. requirements from tasks, number of clients per minute ($\lambda$) or the seed value ($\zeta$) for random number generation (RNG), and is concluded when the percentage of success queries, $q^{S}_t$, falls below a given threshold $\theta$ or when a timeout $H$ is reached. This allows us to speed up the training by terminating unsuccessful or steady episodes quickly.
At every time step $t$, a reward $r_t$ scores the rate of successful queries, see Algorithm~\ref{alg:ASET} at lines 9-10,
where $q^F_t$ is the ratio of ``failed'' queries, i.e., those delivered violating one or more constraints (e.g., out of tolerated delay), and $q^R_t$ is the ratio of queries ``rejected'' by the system for lack of resources, normalized by corresponding time window.  
$\psi$ is a penalty inverse proportional to the episode active time. It ensures that both short and bad action trajectories do not reach higher returns than optimal ones. 
\cancel{By using reward $r_t$,}
\luq{not only that... Next equation needs a good connector with the previous ones in former section, as the TD update algorithm... } \gcdelete{the Q-Values can be estimated iteratively running the following update algorithm:}
\begin{algorithm}[t]
\footnotesize
\caption{\algname\ training procedure}
\label{alg:ASET}
	\begin{algorithmic}[1] 
		\State initialize replay buffer $D = \emptyset$ ring of size $N_r$ and $\phi_{0}$ DQN parameters
		\State initialize training RNG seeds $\zeta \in [0,...,P-1]$
		\For{$\textit{episode}$ k $\in [0,...,M]$}
			\State sample random seed $\zeta \sim \mathcal{U}(P)$
			\State initialize $\pi_{k}(a|s) = \epsilon\; \mathcal{U}(\beta) + (1-\epsilon)\delta(a=\argmax_{a} Q(s,a;\phi_k))$ 
			\For{$\textit{step}$ t $\in [0,...,H]$}
    				\State sample $a_t \sim \pi_{k}(a|s)$
				\State sample $s_{t+1}$ state from edge-network and reward $r_t$
				\If {$\varphi > 1 - (q^F_t + q^R_t)$} $r_t = - (q^F_t + q^R_t)$
                \Else {} $r_t = \psi$
                \EndIf
				\State $D \gets D \cup \{(s_t,a_t,s_{t+1},r_t)\}$ with circular replacement
			\EndFor	
			\If{$q^{S}_t \leq \theta$} $t \gets H$, ends this episode 
			\EndIf
    		\State $\phi_{k,0} \gets \phi_{k}$
			\For{$\textit{gradient descend step}$ g $\in [0,...,G-1]$}
				\State sample batch $B \subset D $ with $idx \sim \mathcal{U}(N_r)$
				\State compute $C_t \gets r_t + \gamma \argmax_{a_{t+1}} Q(s_{t+1}, a(s_{t+1}); \phi_{k})$
				\State compute loss $L = \sum_i ( C_i - Q(s_i,a_i;\phi_k) )^{2}$
				\State update replay network $\phi_{k,g+1} \gets \phi_{k,g} - \alpha\nabla_{\phi_{k,g}}L$ 
			\EndFor
			\State update DQN parameters $\phi_{k+1} \gets \phi_{k,G}$
	    	\EndFor
    	\end{algorithmic}
\end{algorithm}
 Note that DQN network is used to minimize the target loss L, see lines 14-18, by Adam optimizer and $\alpha$ learning rate. It takes gradient steps on the Bellman error objective L, see factor $C$ at line 16 and Eq.~(\ref{eq:qvalue-ite}), concurrently with data collection from the replay buffer~\cite{Lin92self} for an efficient off-line learning.
This is a common hybrid approach to implement Q-Learning~\cite{watkins1989,mnih2015,levine2020}. 
Additionally, we employ an $\epsilon$-greedy exploration policy, see line 5, with parameter $\epsilon$ dynamically updated. \gcdelete{This training protocol is known as off-line reinforcement learning~\cite{levine2020} and complements the on-line one in classical Q-Learning by allowing, among others, the using of efficient stochastic gradient descent (SGD) methods for $\theta$ parameter estimation.}
The architecture of our DQN consists of a stacking of convolutional layers that extracts temporal correlations from the state tensor $\mathbf{S}$. Such feature extraction part is composed of three convolutional layers with 4x4 kernels along the time and feature dimensions, followed by Re-Lu activation and max-pooling. Finally, two linear layers squeeze the dimension from 256 to as many outputs as different static policies $\beta$. 

\vspace{-0.12cm}
\section{Performance evaluation}
\label{sec:validation}

We evaluate \algname\ using a prototype implementation of an edge inference system that will be released upon acceptance of the paper.
We first use our prototype to run small scale experiments with the aim of profiling some representative models and their variants (results not shown).
Then we use such profiling data to run large scale experiments on a simulated setup, comparing the performance of \algname\ to those of static scheduling policies.
\vspace{-0.12cm}
\subsection{Evaluation settings}
\label{subsec:setup}

\noindent
\textbf{System Prototype.}
Our prototype implements the edge inference system functionalities described in Section~\ref{sec:design}.
On each cluster, a \textit{Master} deploys workers and routes streams between them and remote clients; each \textit{Worker} runs in a Docker container and implements a pipeline that processes queries in FIFO order from different streams, based on the model variant batch size; a \textit{Monitoring} agent on each cluster collects stats from model variants usage and their performance, used \textit{(i)} to build a catalog of model variants and \textit{(ii)} to provide each \textit{Scheduler} with aggregated observations on the system state.
We use such a prototype to profile variants of pre-trained inference models with respect to their resource usage and performance (see below).

\noindent
\textbf{Simulation Setup.}
To evaluate our approach on a large scale, we set up a simulated environment where each worker simulates the inference task based on the profiling information available for its model variant. Therefore, empty responses are generated for each batch of queries after simulating a processing delay (based on a normal distribution). Additionally, we simulate network delay between stream sources and destination clusters (see below for considerations on the network topologies), as well as the transmission delay. 
Apart from simulated workers, other system components 
are deployed using their prototype implementation. Therefore, the system operates on a realistic timescale.

\noindent
\textbf{Network topology.}
We leverage the network topology of a large ISP to assess scheduling performance under realistic settings. Specifically, our simulated environment is a cloud-to-edge topology with clusters of different sizes deployed hierarchically. To preserve ISP confidentiality, we only report a high-level summary of topology, latency, and hardware distribution characteristics.
Similarly to the tiered topologies from~\cite{tong2016hierarchical, ceselli2017mobile},
our topology can  
provide clusters with computation capabilities at different layers: network access (e.g., antennas, home gateway), central offices (multiple payers), operator data center, and remote cloud (third parties). Specifically, we focus on three scenarios: \textit{(i)} \emph{dc-cloud}, where resources are deployed at ISP data center and remote cloud only; \textit{(ii)} \emph{co-dc-cloud}, where resources are deployed at central offices, operator data center and remote cloud; \textit{(iii)} \emph{full-edge} topology, where clusters are deployed at all layers previously mentioned. Note that we limit the simulations to the topology serving 1,000 antennas from the full ISP topology, and appropriately scale resources (see below).
For the evaluation, we assume a 5G radio access technology with antennas deployed similarly to LTE.
Network/transmission delays range from few milliseconds, to reach the eNodeBs behind the antennas, to the order of ten milliseconds for central offices and ISP data centers, and few tens of milliseconds for the remote cloud.


\begin{table}[t]
\footnotesize
    \centering
    \caption{Characteristics of reference applications~\cite{chen2017empirical, cartas2019reality}.}
    \label{tab:apps-metrics}
    
    \begin{tabular}{l|r|r|r|r}
        \toprule
            \multicolumn{1}{c}{\centering \textbf{Edge app}} 
            & \multicolumn{1}{p{1.1cm}}{\centering \textbf{Tolerated \\ delay}} 
            & \multicolumn{1}{p{1.2cm}}{\centering \textbf{Frame \\ rate}} 
            & \multicolumn{1}{p{1.0cm}}{\centering \textbf{Streams \\ duration}} 
            &\multicolumn{1}{p{1cm}}{\centering \textbf{Required \\ accuracy}}
        \\
        \midrule
            Pool
            & 95 ms
            & 5 FPS
            & 5-10 s
            & 10 mAP
        \\
            Workout Assistant
            & 300 ms
            & 2 FPS
            & 90 s
            & 10 mAP
        \\
            Ping-pong
            & 150 ms
            & 15-20 FPS
            & 20-40 s
            & 15 mAP
        \\
            Face Assistant
            & 370 ms
            & 5 FPS
            & 1-5 s
            & 30 mAP
        \\
            Lego/Draw/Sandwich
            & 600 ms
            & 10-15 FPS
            & 60 s
            & 25 mAP
        \\
            Gaming
            & 20-30 ms
            & 25 FPS
            & 10-30 m
            & 35 mAP
        \\
            Connected Cars
            & 150 ms
            & 10-15 FPS
            & 15-30 m
            & 40 mAP
        \\
            Tele-Robots
            & 25-35 ms
            & 10 FPS
            & 5 m
            & 40 mAP
        \\
            Remote-driving
            & 20-30 ms
            & 20 FPS
            & 15-30 m
            & 50 mAP
        \\
            Interactive AR/VR
            & 30-50 ms
            & 25 FPS
            & 30-60 s
            & 35 mAP
        \\
        \bottomrule
    \end{tabular}
\end{table}

\noindent
\textbf{Requests workload.}
Requests are generated following a Poisson distribution. Each generator runs on average $\lambda$ clients per minute querying the scheduler of a given geographical area (antenna). Once spawned, each client requests for processing a stream featuring randomized characteristics in terms of frame rate, 
required end-to-end latency,
required model accuracy,
frame sizes, 
stream duration
. To capture realistic queries characteristics, we modeled 
metrics of generated streams according to the reference edge applications in Table~\ref{tab:apps-metrics}. In our settings, a generator with $\lambda$ = 60 brings a load of almost 1000 queries per second on the serving antenna.

\noindent
\textbf{Computing clusters and model variant.}
We assume a given reference hardware distribution across clusters, with computing capabilities increasing from the access network to the cloud. Specifically, the access network can be equipped with an 8-16 cores machine, 16 GB of memory and a small TPU, central offices can host in the order of tens of servers (32-64 CPUs, 128-256 GB, and few GPUs), ISP data centers can host hundreds of servers, while for the centralized cloud we assume unlimited resources.
In our evaluation, we focus on DNN models for the object detection task, as it is one of the most challenging and computation-intensive inference service~\cite{jiao2019survey, srivastava2018performance}. 
Using our prototype we profile MobileNet-SSD, Yolo-v3, and Tinyyolo-v2 models~\cite{liu2016ssd, redmon2016you}, with CPU and GPU variants on different batch sizes, scaling on allocated resources and number of replicas. Such a set of results is not shown for lack of space.
We use profiled information to run our simulations on top of the three topologies described above.
On each cluster, workers have been scaled on the number of replicas up to resource saturation.

\begin{figure}[t]
    \centering
    \begin{subfigure}[t]{\columnwidth}
        \centering
        \includegraphics[clip= true, width=\linewidth, trim=0in 0.0in 0in 0in]{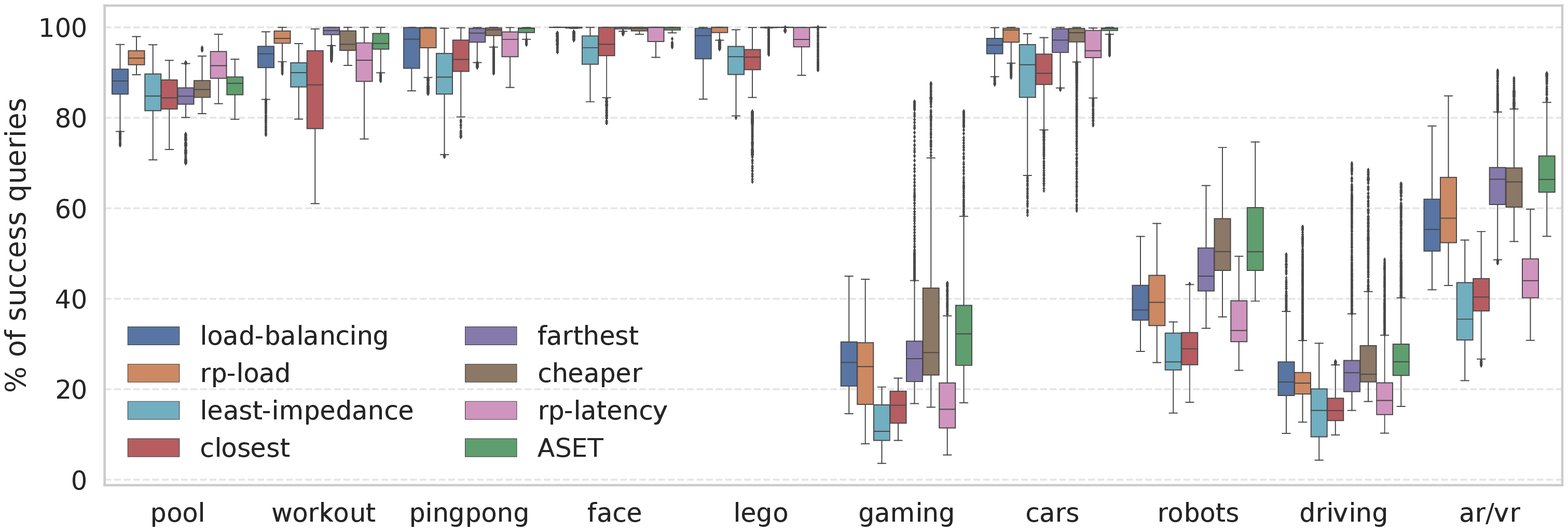}
        \vspace{-6mm}
        \caption{Average values for clients rate $\lambda$ = 60}
        \label{subfig:apps-policies-60}
    \end{subfigure}\\
    \begin{subfigure}[t]{\columnwidth}
        \centering
        \includegraphics[clip= true, width=\linewidth, trim=0in 0in 0in 0in 4.2in]{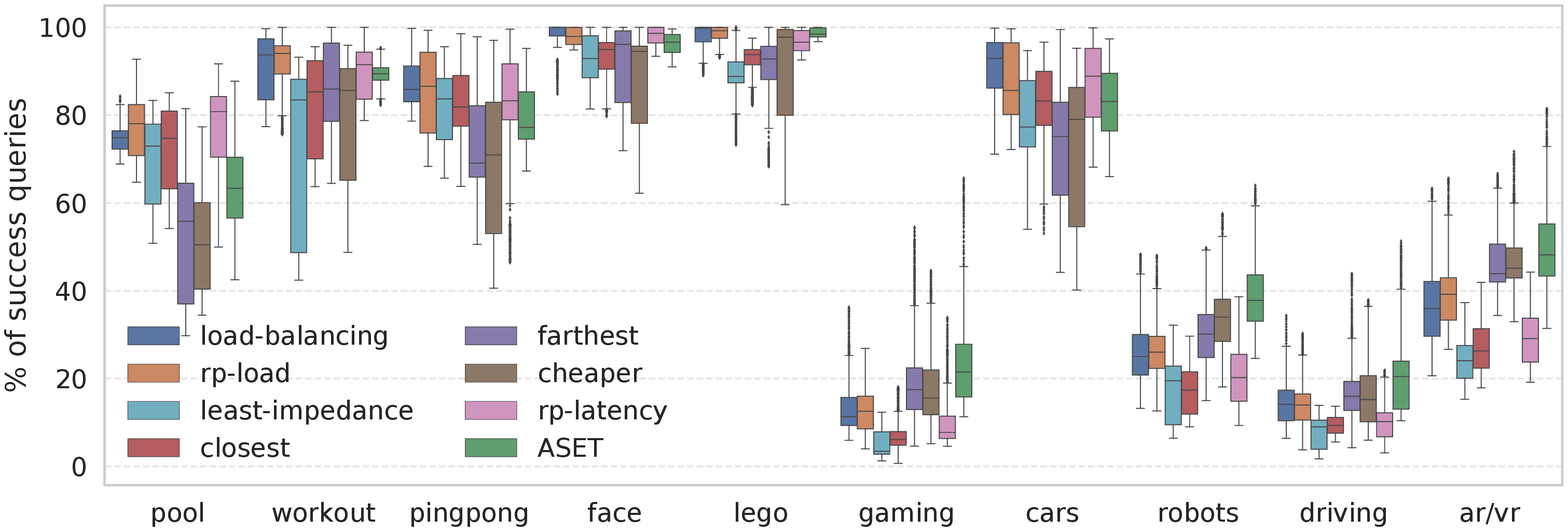}
        \vspace{-6mm}
        \caption{Average values for episodes with dynamic clients rates.}
        \label{subfig:apps-policies-changing}
    \end{subfigure}%
    \caption{Success percentage for different apps on the full-edge topology.}
    \label{fig:apps-policies}
\end{figure}

\vspace{-0.20cm}
\subsection{Experimental Results}
\label{subsec:results}

We compare the performance of the baseline policies described in Section~\ref{subsec:policies} distinguishing results for different applications from Table~\ref{tab:apps-metrics}. As a performance metric we consider the percentage of queries that are successfully processed by the system satisfying the application QoS requirements.
Figure~\ref{subfig:apps-policies-60} shows results of multiple runs with $\lambda$ = 60. Results suggest that there is no one-size-fits-all policy, as various applications may benefit differently from each policy. Varying the rate of stream requests on the antenna (Figure~\ref{subfig:apps-policies-changing}) may further increase the uncertainty of relying on a single policy.
In the following, we compare the performance of the \algname\ RL scheduling approach with the performance of static policies, evaluating the benefits it can introduce in the various scenarios. We trained three different versions of \algname\ (one for each topology). In particular, we sample the state using a time window $T$ = 25 seconds, and we experimentally chose an episode timeout of 8 minutes to avoid steady states in the network. Despite we evaluate on multiple clients rate, our agent has been trained only on episodes with $\lambda$ = 60.

\begin{figure}[t]
    \centering
    \begin{subfigure}[t]{0.49\columnwidth}
        \centering
        \includegraphics[clip= true, width=\linewidth, trim=0in 0in 0in 0in]{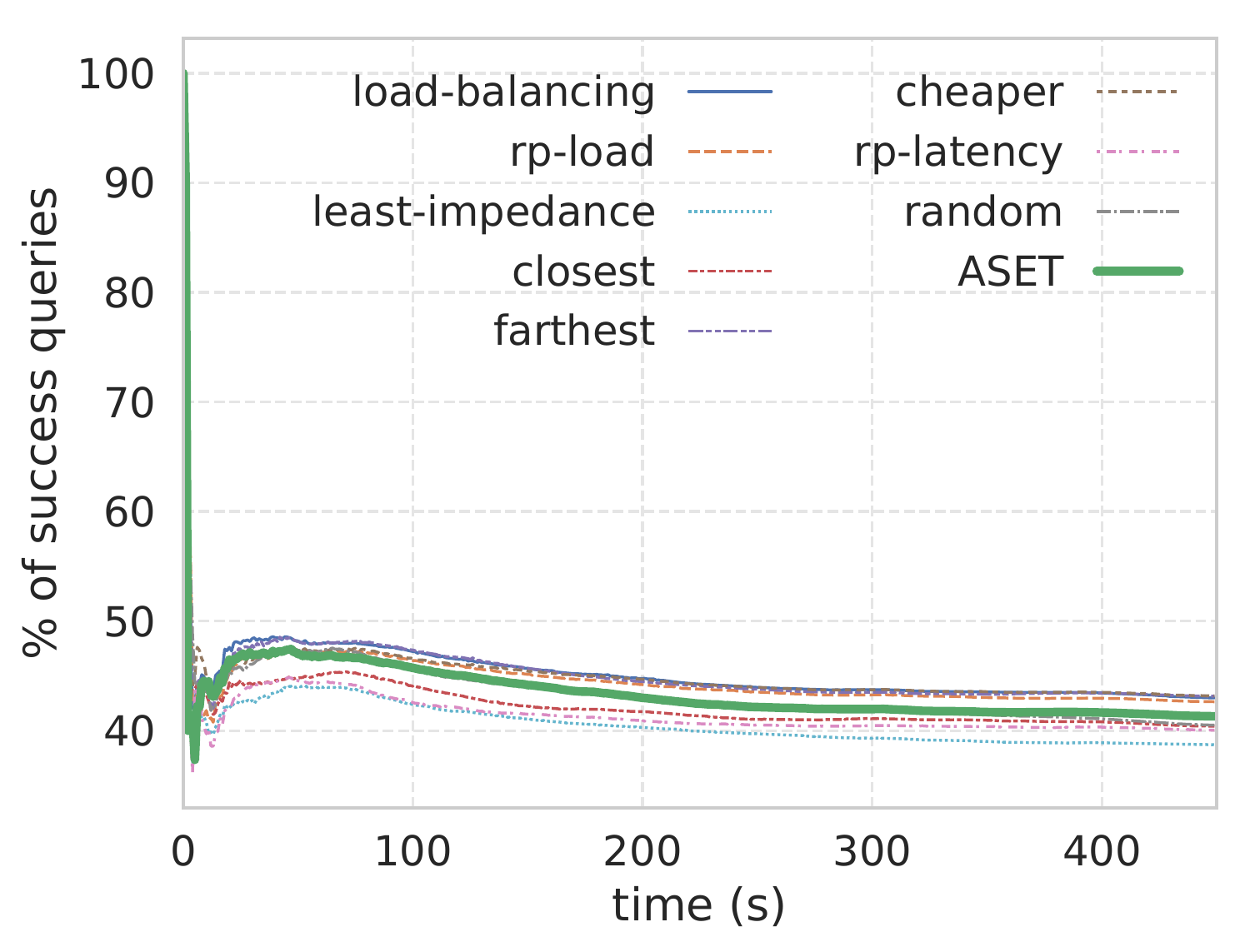}
        \vspace{-6mm}
        \caption{Time avg on dc-cloud for $\lambda$ = 60.}
        \label{subfig:cloud-success}
    \end{subfigure}
    \begin{subfigure}[t]{0.49\columnwidth}
        \centering
        \includegraphics[clip= true, width=\linewidth, trim=0in 0in 0in 0in]{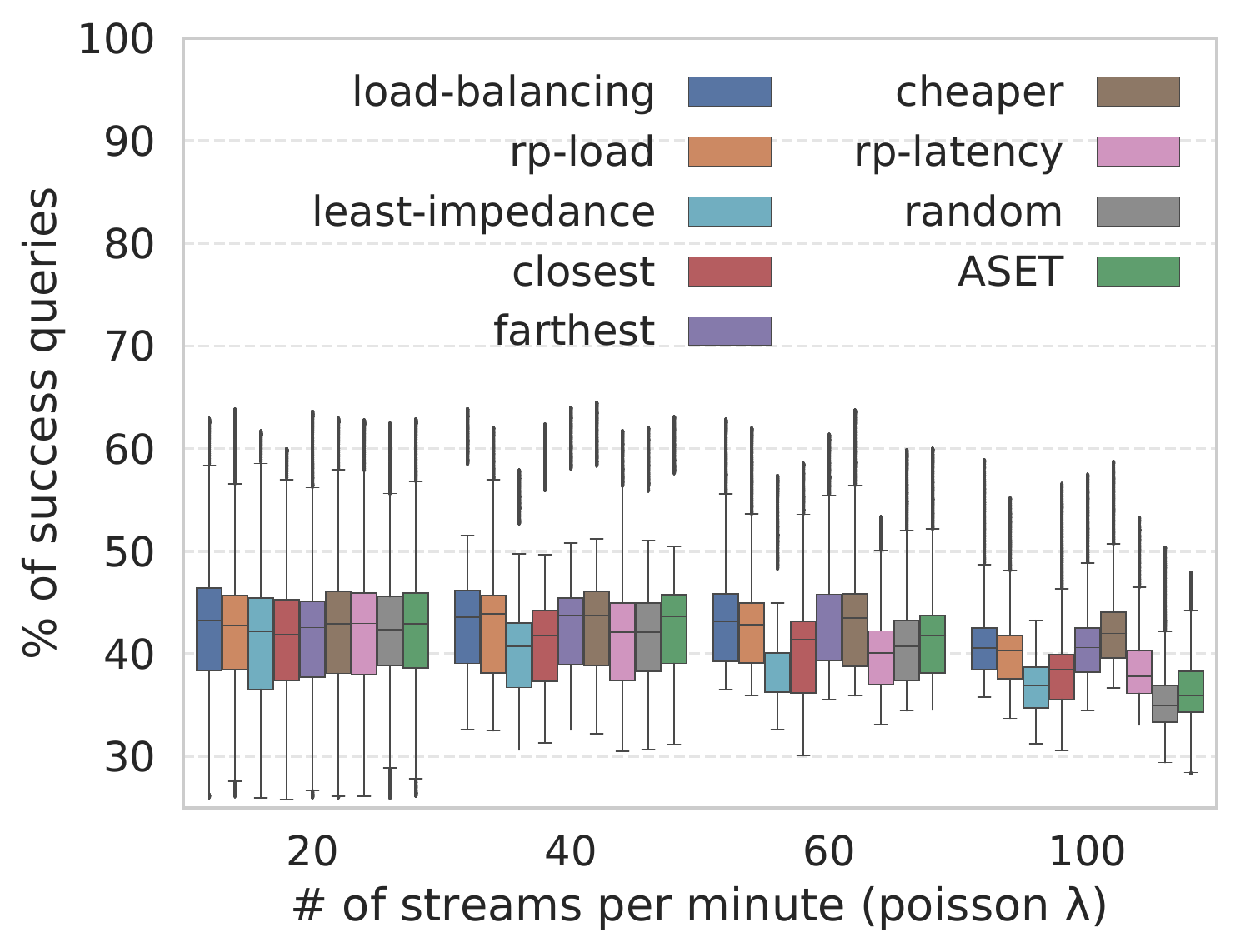}
        \vspace{-6mm}
        \caption{Different clients rate on dc-cloud.}
        \label{subfig:cloud-lambdas}
    \end{subfigure}\\
    \begin{subfigure}[t]{0.49\columnwidth}
        \centering
        \includegraphics[clip= true, width=\linewidth, trim=0in 0in 0in 0in]{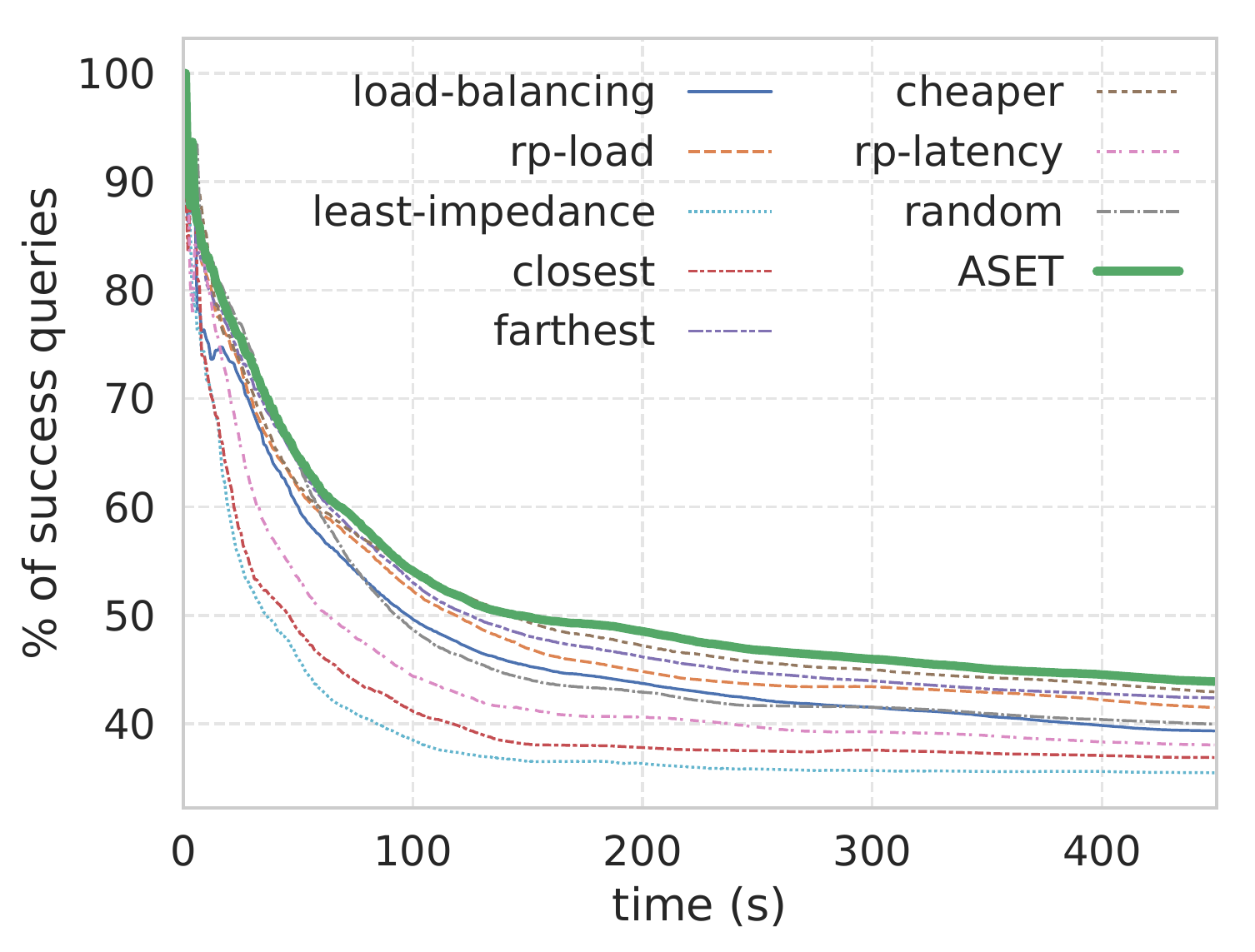}
        \vspace{-6mm}
        \caption{Time avg on co-dc-cloud for $\lambda$ = 60.}
        \label{subfig:co-success}
    \end{subfigure}
    \begin{subfigure}[t]{0.49\columnwidth}
        \centering
        \includegraphics[clip= true, width=\linewidth, trim=0in 0in 0in 0in]{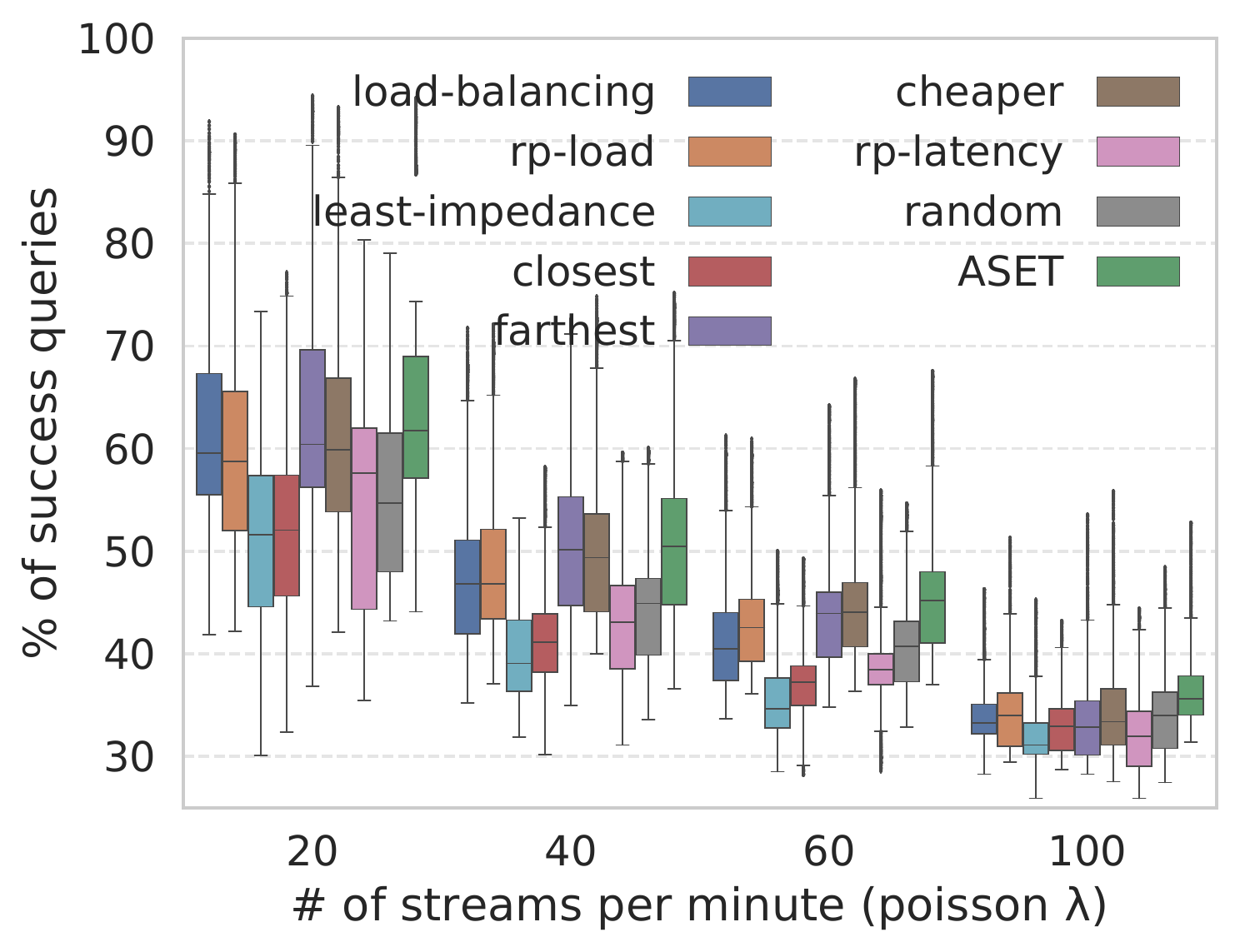}
        \vspace{-6mm}
        \caption{Different clients rate on co-dc-cloud.}
        \label{subfig:co-lambdas}
    \end{subfigure}
    \caption{Performance of \algname\ compared with static policies for (ab) the dc-cloud topology and (cd) the co-dc-cloud topology.}
    \label{fig:cloud-topology}
\end{figure}

\noindent
\textbf{Cloud deployment.}
When all the available resources are located in a few centralized clusters, the various static policies have small differences in performance and a dynamic approach has little room for improvement.
Results for the dc-cloud topology are shown in Figures~\ref{fig:cloud-topology}ab. In particular, Figure~\ref{subfig:cloud-success} plots, for every moment of the simulation (time axis), the percentage of queries that are handled successfully, averaging multiple runs with different workloads. The graph shows that, for this topology, \algname\ does not improve over static policies, and it even performs worse for higher lambdas (Figure~\ref{subfig:cloud-lambdas}). 
Figures~\ref{fig:cloud-topology}cd shows that moving some resources to Central Offices (co-dc-cloud topology) makes a huge difference: in general, all the policies achieve a higher success ratio on this configuration (Figure~\ref{subfig:co-success}), as they can exploit the additional lower latency spots, and the higher level of distribution gives to \algname\ a certain margin of improvement. Figure~\ref{subfig:co-lambdas} shows that \algname\ introduces some improvement over all the baselines for every lambda, despite being trained only for $\lambda$ = 60.

\begin{figure}[t]
    \centering
    \begin{subfigure}[t]{0.49\columnwidth}
        \centering
        \includegraphics[clip= true, width=\linewidth, trim=0in 0in 0in 0in]{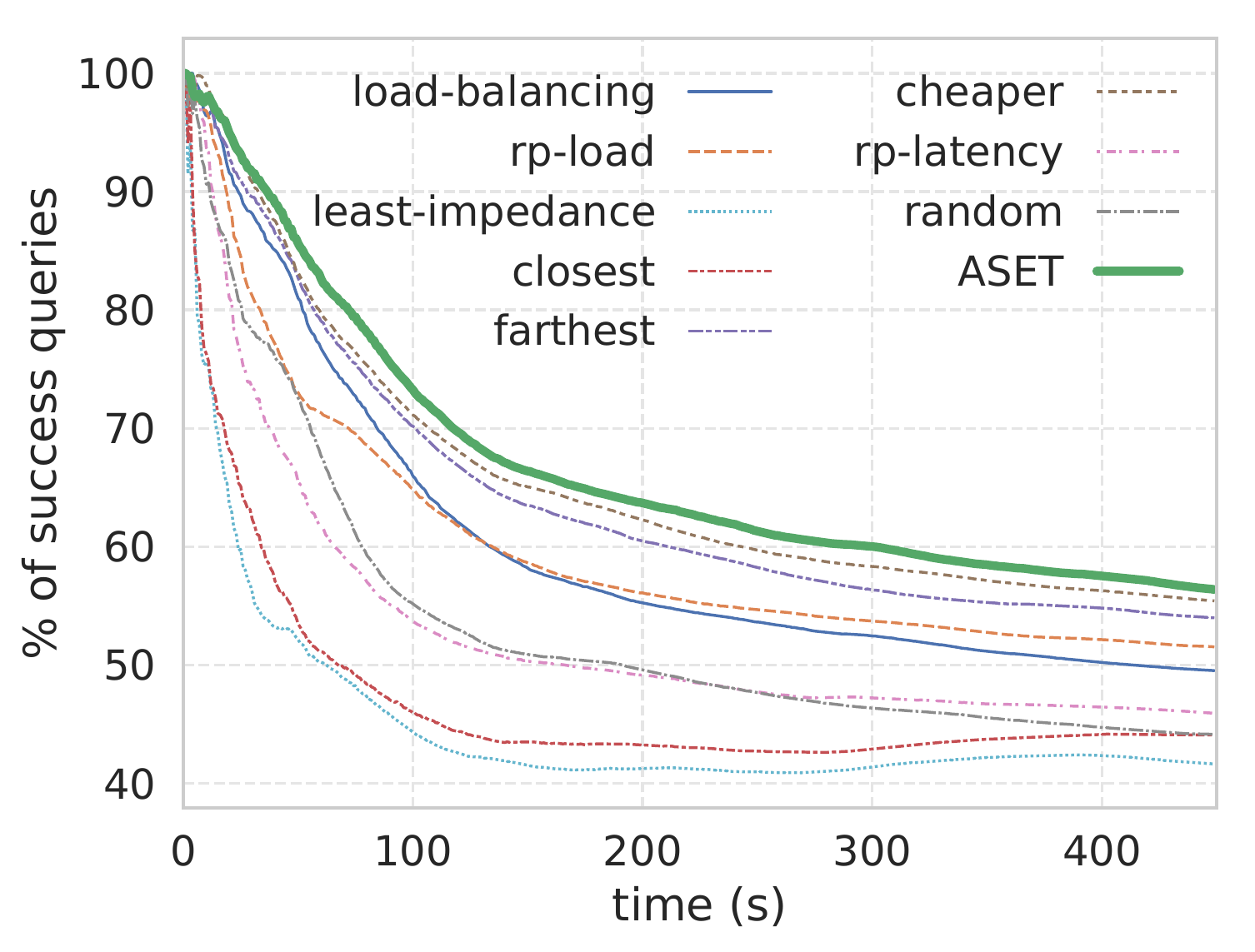}
        \vspace{-6mm}
        \caption{Queries handled successfully.}
        \label{subfig:edge-success}
    \end{subfigure}
    \begin{subfigure}[t]{0.49\columnwidth}
        \centering
        \includegraphics[clip= true, width=\linewidth, trim=0in 0in 0in 0in]{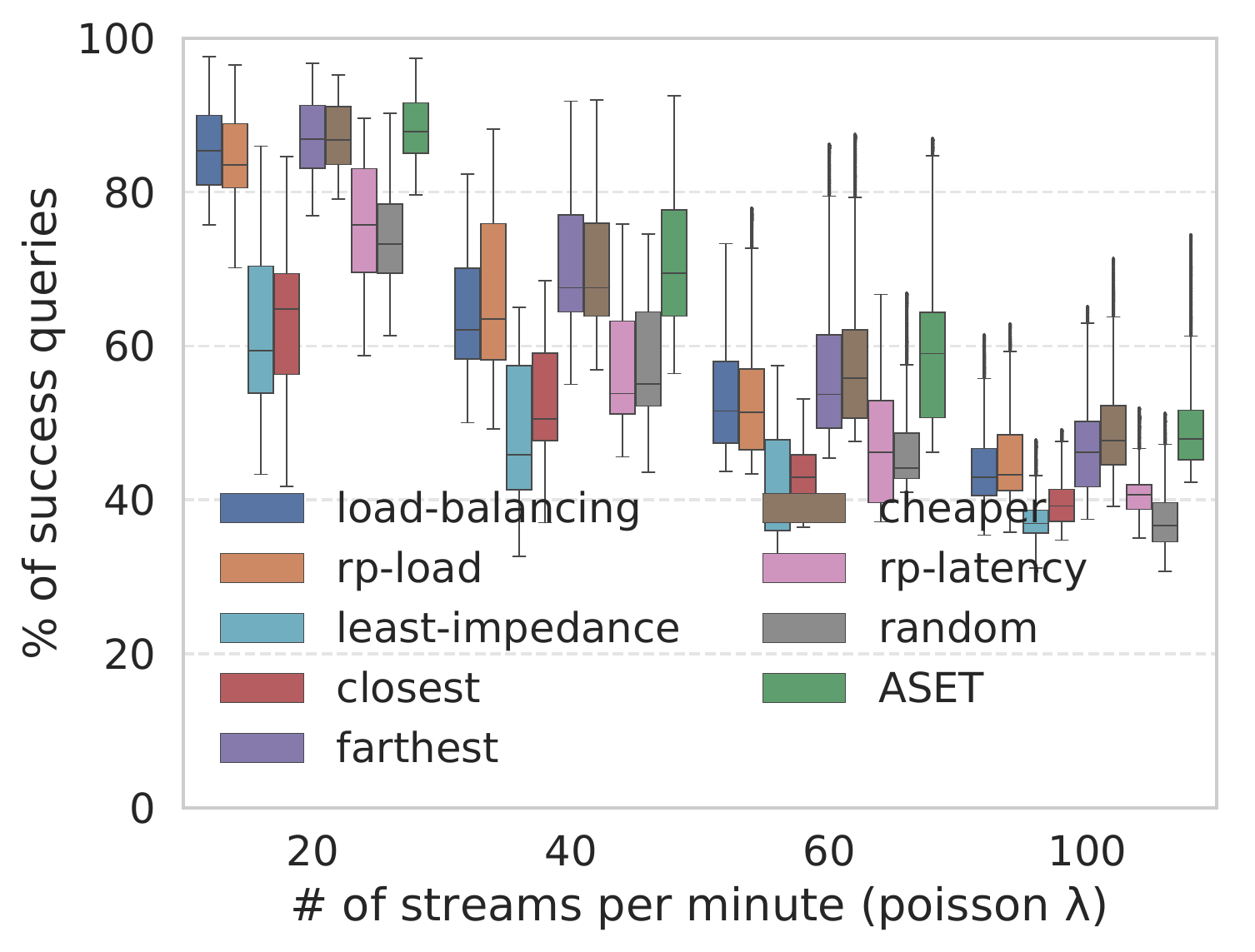}
        \vspace{-6mm}
        \caption{Different clients rate on full-edge.}
        \label{subfig:edge-lambdas}
    \end{subfigure}\\
    \begin{subfigure}[t]{0.49\columnwidth}
        \centering
        \includegraphics[clip= true, width=\linewidth, trim=0in 0in 0in 0in]{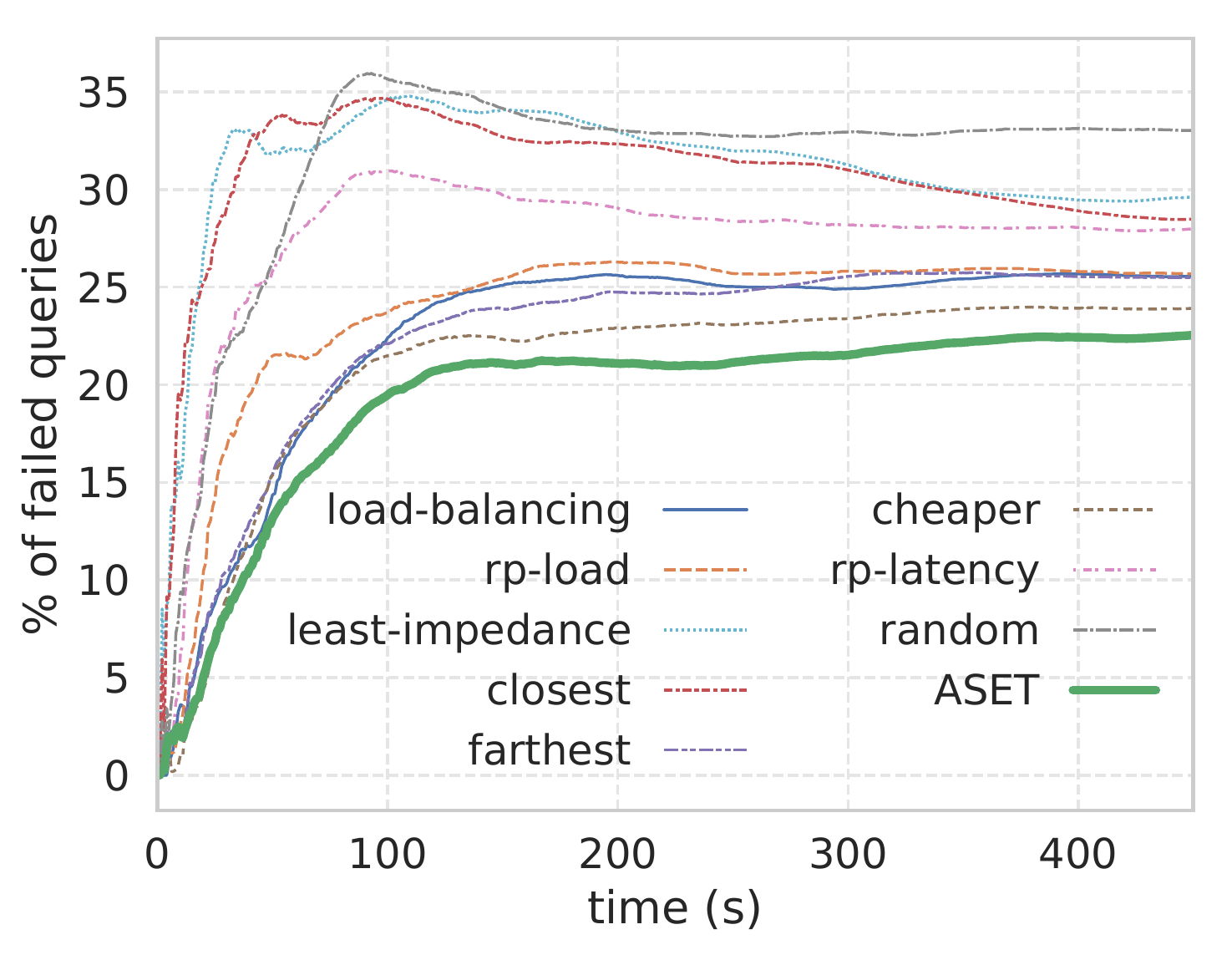}
        \vspace{-6mm}
        \caption{Queries delivered with QoS violations.}
        \label{subfig:edge-failures}
    \end{subfigure}
    \begin{subfigure}[t]{0.49\columnwidth}
        \centering
        \includegraphics[clip= true, width=\linewidth, trim=0in 0in 0in 0in]{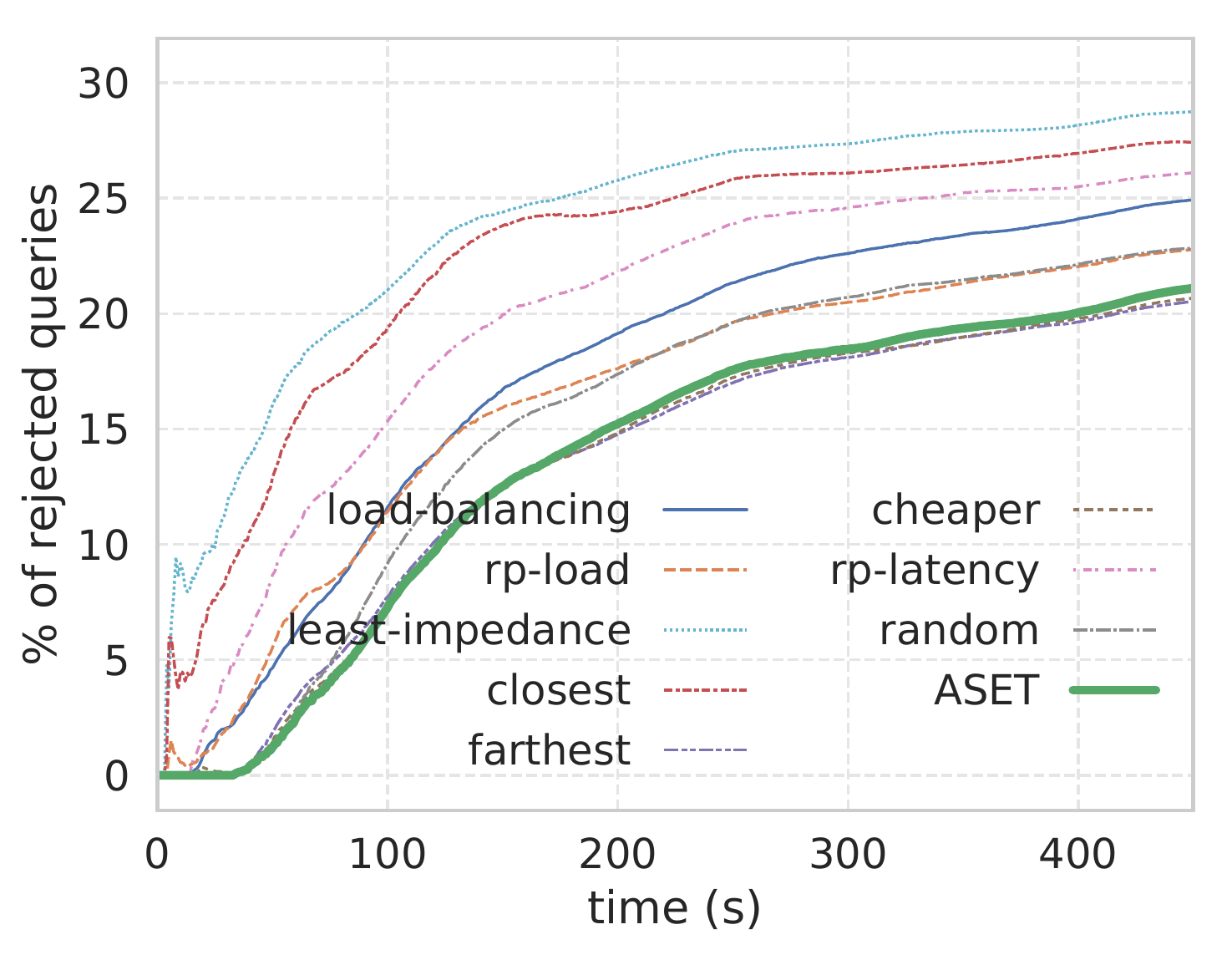}
        \vspace{-6mm}
        \caption{Queries rejected for lack of resources. \luq{Is the time axis representing the maximum delay requested by the query?}\gabriele{I have better explained in the main text what the time axis stands for (is the simulation time)}}
        \label{subfig:edge-rejections}
    \end{subfigure}
    \caption{Performance of \algname\ compared with static policies for the full-edge topology. (a) (c) and (d) show averages of multiple runs with $\lambda$ = 60. 
    }
    \label{fig:edge-topology}
\end{figure}

\noindent
\textbf{Edge deployment.}
The results so far suggest that a good distribution of computing resources is a key factor to improve against static scheduling policies. As shown in Figure~\ref{fig:edge-topology}, the benefits of using a dynamic scheduling approach become more concrete in a full-edge topology, where resources are better distributed on multiple smaller clusters in different locations. 
In fact,
Figure~\ref{subfig:edge-success} shows that the dynamic approach of \algname\ is able to achieve a constant improvement over any static policy, with a higher success ratio over time. In particular, Figures~\ref{fig:edge-topology}cd show that, while maintaining the same rejection rate as the best static-policy, \algname\ effectively reduces the number of queries that are handled violating one or more QoS requirements.
Moreover, Figure~\ref{subfig:edge-lambdas} shows that an \algname\ agent trained only for $\lambda$ = 60 can also generalize on different requests rate, even supporting a load of more than 1600 queries per second ($\lambda$ = 100) on a single antenna.

\begin{figure}[t]
    \centering
    \begin{subfigure}[t]{0.50\columnwidth}
        \centering
        \includegraphics[clip= true, width=\linewidth, trim=0in 0in 0in 0in]{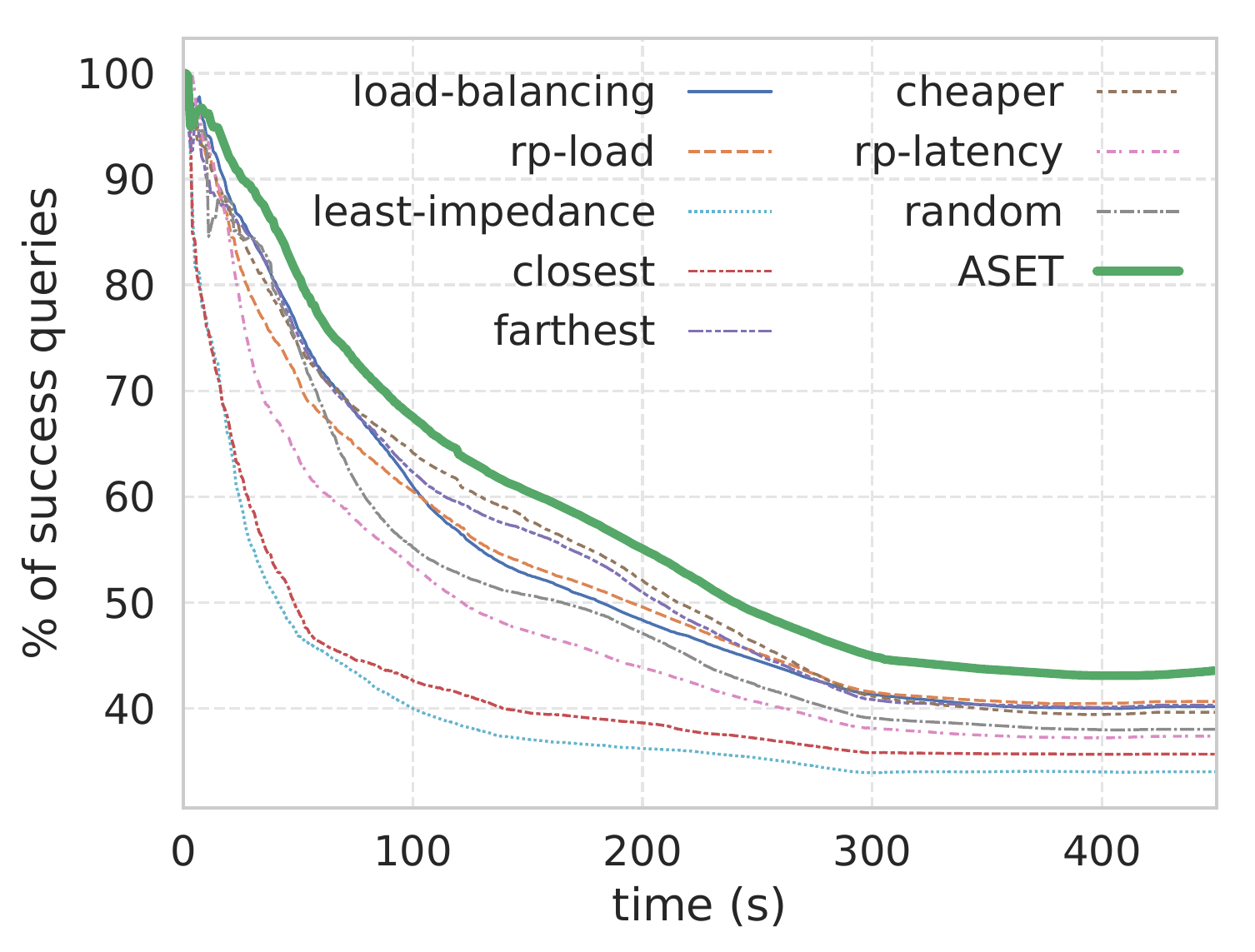}
        \vspace{-6mm}
        \caption{Burst variation of $\lambda$ from 20 to 100.}
        \label{subfig:dynamic-1}
    \end{subfigure}%
    \begin{subfigure}[t]{0.50\columnwidth}
        \centering
        \includegraphics[clip= true, width=\linewidth, trim=0in 0in 0in 0in]{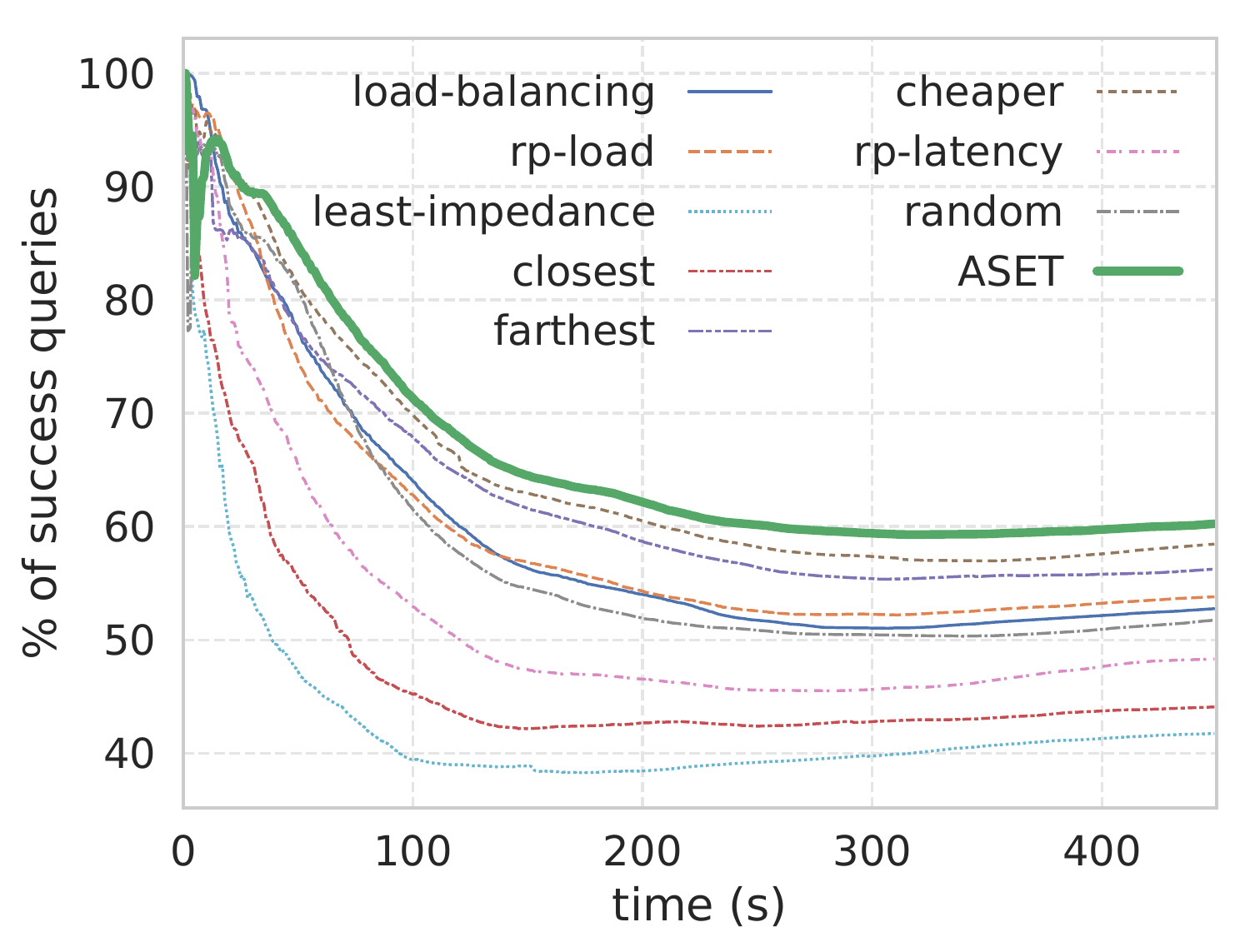}
        \vspace{-6mm}
        \caption{Steady variation of $\lambda$ between 60 and 20.}
        \label{subfig:dynamic-2}
    \end{subfigure}%
    \caption{Performance of \algname\ varying the requests rate over time with two different load variation patterns (full-edge topology).\luq{Is missing same curves to compare with a different state space. }}
    \label{fig:dynamic-lambdas}
\end{figure}

\noindent
\textbf{Dynamic input rate.}
We have performed some additional experiments to evaluate how the system behaves in dynamic situations where the requests rate varies over time. For this purpose, we have set up some dynamic runs where the lambda value changes every 150 seconds: a first pattern simulates a particularly fast variation with values of 20, 60, and 100 clients per minute (Figure~\ref{subfig:dynamic-1}); a different pattern simulates a more steady scenario where the requests rate first moves from 60 to 40 clients per minute, then drops to 20, and finally slowly goes back to 60 (Figure~\ref{subfig:dynamic-2}).
Similar to previous plots, the outcomes for this set of experiments are shown averaging values over time for multiple runs (Figure~\ref{fig:dynamic-lambdas}).
Results in both figures show that having a dynamic requests arrival even introduces a bigger margin for improvement that \algname\ effectively exploits reaching the highest percentage of queries handled successfully. This appears particularly evident in the case where the variation between client arrivals is faster and bigger (Figure~\ref{subfig:dynamic-1}). This result suggests that, while some of the static policies may achieve decent performance when the system load is stable, they struggle on more dynamic scenarios. In such situations, an adaptive algorithm such as \algname\ is more suitable as it can learn how to best optimize the system under different conditions. Moreover, results suggest that \algname\ training generalizes enough as the algorithm performs 
well under previously unseen dynamic conditions.

\begin{figure}[t]
    \centering
    \begin{subfigure}[t]{0.50\columnwidth}
        \centering
        \includegraphics[clip= true, width=\linewidth, trim=0in 0in 0in 0in]{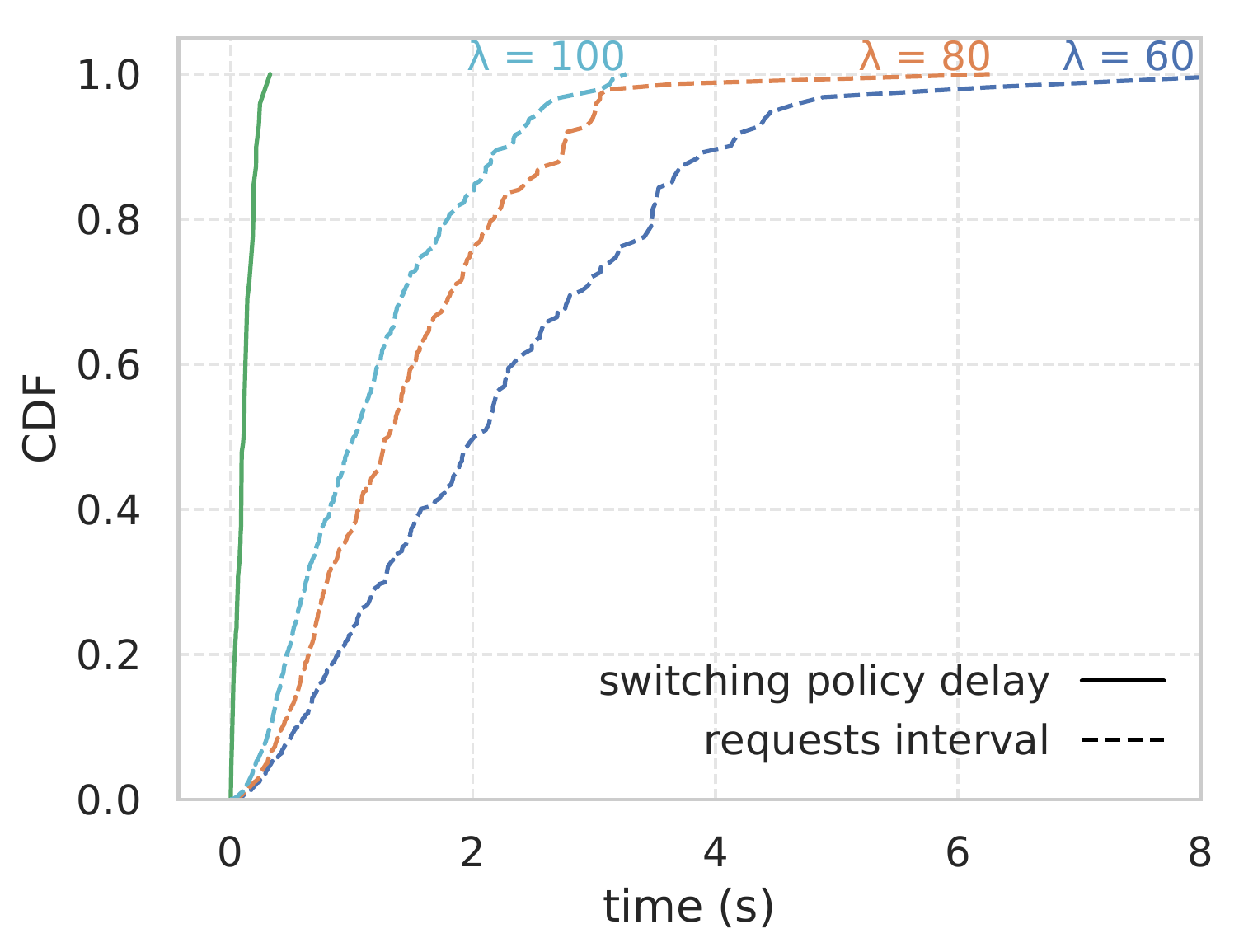}
        \vspace{-6mm}
        \caption{Cumulative distribution of delay.\luq{But how is possible less than 5 seconds? that is our time step for feature window analysis.}\gabriele{only considers the step where the agent infer a new policy: it is the interval from building the state to set the new policy.}}
        \label{subfig:cdf-arrivals}
    \end{subfigure}%
    \begin{subfigure}[t]{0.50\columnwidth}
        \centering
        \includegraphics[clip= true, width=\linewidth, trim=0in 0in 0in 0in]{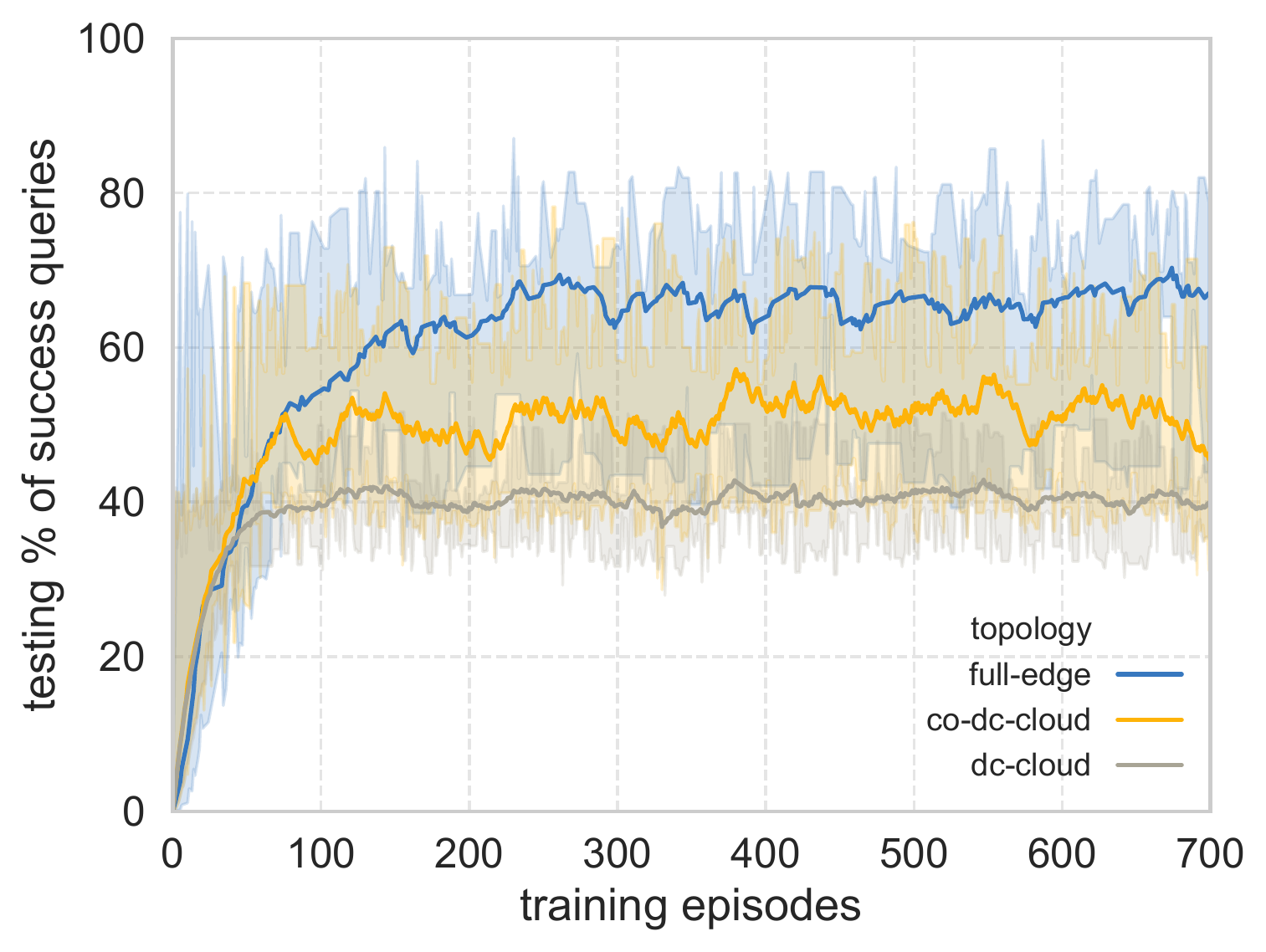}
        \vspace{-6mm}
        \caption{Learning curve.}
        \label{subfig:applicability-training}
    \end{subfigure}%
    \caption{(a) Delay for switching policy compared with requests arrival intervals. (b) Learning curve while training \algname\ on different topologies.}
    \label{fig:applicability}
\end{figure}

\noindent
\textbf{Training and applicability.}
Figure~\ref{subfig:cdf-arrivals} shows the cumulative distribution of the time needed by \algname\ to infer a switching decision from the current policy to the one that is best suitable for the current system conditions (non-dashed line). The switching delay is compared with distributions of the intervals between subsequent requests for different lambdas. As shown, even for very large client loads (100 clients connected to the antenna), the time interval between two stream arrivals is typically in the scale of seconds or hundreds of milliseconds, while the delay for switching between policies is one order of magnitude smaller.
Finally, Figure~\ref{subfig:applicability-training} shows the learning curve of \algname\ for different topologies on continuous stream request arrivals. 
The figure shows that \algname\ quickly reaches a certain level of performance in the first training iterations (before 100 episodes) 
independently from the topology complexity, leaving room for extra improvements to the subsequent episodes based on the margin left by the topology itself. 

\vspace{-0.10cm}
\section{Conclusions}
\label{sec:conclusions}

This paper proposes \algname, an adaptive algorithm based on Reinforcement Learning for scheduling inference workloads at the network edge. \algname\ solves the problem of exploiting scattered clusters of resources to serve inference queries from multiple edge applications (e.g., AR/VR, cognitive assistance). We model an edge inference system where queries from different access networks are processed across a multitude of distributed processing locations. The constrained nature of the edge network introduces a trade-off between network delay and processing time based on the various available DNN models. In such a scenario, \algname\ optimizes the binding between inference stream requests and available DL models across the network, maximizing the throughput and ensuring that any requirement in terms of inference accuracy and end-to-end delay is satisfied.
We evaluated our approach over the realistic network topology of a large ISP and considering a heterogeneous pool of edge applications. Our findings show that \algname\ effectively improves the performance compared to static policies when resources are deployed across the whole edge-cloud infrastructure.

\bibliographystyle{IEEEtran}
\bibliography{paper}

\end{document}